\documentclass{article} %
\usepackage{placeins}
\usepackage{iclr2027_conference,times}

\usepackage{amsmath,amsfonts,bm}

\def\eqref#1{equation~\ref{#1}}

\def\1{\bm{1}}

\DeclareMathAlphabet{\mathsfit}{\encodingdefault}{\sfdefault}{m}{sl}
\SetMathAlphabet{\mathsfit}{bold}{\encodingdefault}{\sfdefault}{bx}{n}

\usepackage{hyperref}
\usepackage{url}
\usepackage[utf8]{inputenc}
\usepackage[T1]{fontenc}
\usepackage{booktabs}
\usepackage{amsfonts,amsmath,amssymb}
\usepackage{nicefrac}
\usepackage{microtype}
\usepackage{xcolor}
\usepackage{graphicx}
\usepackage{subcaption}
\usepackage{multirow}
\usepackage{tabularx}
\usepackage{rotating}
\usepackage{enumitem}
\newcommand{\corrTapeN}{22}

\newcommand{\rhoReadingJudgment}{0.23}
\newcommand{\pReadingJudgment}{= 0.27}

\newcommand{\rhoJudgmentMeasurement}{0.61}

\newcommand{\rhoFoneJudgment}{0.77}

\newcommand{\rhoFoneReading}{0.26}
\newcommand{\pFoneReading}{= 0.19}
\newcommand{\rhoFoneMeasurement}{0.77}

\newcommand{\rhoFoneDoc}{0.69}
\newcommand{\pFoneDoc}{< 0.001}

\newcommand{\nFields}{24}

\newcommand{\docEngineAdq}{0.71}
\newcommand{\docEngineInadq}{0.50}
\newcommand{\docEngineDeltaMin}{0.14}
\newcommand{\docEngineDeltaMax}{0.26}

\newcommand{\tapeLeanShareInadq}{88}
\newcommand{\tapeSaidTilted}{75}

\newcommand{\nModels}{26}
\newcommand{\nFrontier}{7}

\newcommand{\tapeRollRisesModels}{23}

\newcommand{\tapeRollRisesFrontier}{5}

\newcommand{\misTapeRate}{75}
\newcommand{\misTapeShare}{97}

\newcommand{\misSlopeShare}{74}

\newcommand{\misRailShare}{63}
\newcommand{\misRejMin}{89}
\newcommand{\misRejMax}{99.6}

\newcommand{\orderN}{150}
\newcommand{\orderModels}{9}
\newcommand{\orderFrontierModels}{4}

\newcommand{\humanFone}{0.59}
\newcommand{\humanFoneLo}{0.50}
\newcommand{\humanFoneHi}{0.67}

\newcommand{\humanWage}{35.91}
\newcommand{\humanWageLo}{10}
\newcommand{\humanCostLo}{11}
\newcommand{\humanCostHi}{165}
\newcommand{\humanItems}{1,378}

\graphicspath{{figs/}}
\hypersetup{colorlinks=true,linkcolor=blue!50!black,citecolor=blue!50!black,urlcolor=blue!50!black}   %

\newif\ifarxiv
\arxivtrue
\ifdefined\setarxiv\arxivtrue\fi
\ifarxiv\newcommand{\arxivonly}[1]{#1}\newcommand{\reviewonly}[1]{}\else\newcommand{\arxivonly}[1]{}\newcommand{\reviewonly}[1]{#1}\fi
\newcommand{\dataurl}{\url{https://huggingface.co/datasets/relantic/realitybench}}

\newcommand{\bench}{RealityBench}
\newcommand{\env}[1]{\texttt{#1}}

\title{\bench: Simulated Environments to Evaluate a Vision Language Model's Ability to Document Human Physical Work}

\author{Andrew Perrault, Anmol Takiar, Dominik Winecki, Arnab Nandi \\
Relantic, Inc.\\
Columbus, OH, USA \\
\texttt{\{andrew,anmol,dominik,arnab\}@relantic.com} \\
}

\ifarxiv\iclrfinalcopy\fi
\begin{document}

\maketitle
\ifarxiv\lhead{Preprint}\fi

\begin{abstract}
Augmenting worker capabilities in the physical world through guidance has been a decades-long goal for AI agents. One key supporting capability for agents is to be able to understand what the worker is doing, seeing through the human’s camera or smart glasses, and assist them in documentation. Despite impressive advances in vision language models, their capabilities in these settings remain uncertain and are challenging to benchmark because most physical work data is proprietary, and involves too much variability to study in a principled manner. 
Unlike structured settings such as kitchens and warehouses that have predictable lighting, geometry, and objects, field-facing tasks such as rooftop solar installation and vehicle repair are unconstrained and high variance. 
Here, we model real-world physical work scenarios in Blender across eleven distinct environments, incorporating variants in both image acceptability and documentation.
The acceptability task asks whether a particular view would be suitable proof-of-work and documentation asks whether an agent can understand the task-relevant state of the environment from that view. Across \nModels{} vision-language models, we find that top models largely solve documentation when the answer is text in frame, but are weak at reading measurements and at judging small parts and spatial relations, such as whether a tape measure leans. Acceptability is harder to pin down because human raters disagree with each other. Measured against a scorer that agrees with the raters' majority about as often as a single rater does, the best models cannot be distinguished from a single non-author rater, at a cost within the range of a human pass. 
In the three environments where we can trace acceptability errors, most wrong rejections come from misreading the scene rather than the requirement. On the three documentation questions for which we have real images, model rankings on rendered frames transfer to the real images. The benchmark is available at \dataurl.
\end{abstract}

\section{Introduction}
\begin{figure}[t]\centering\includegraphics[width=0.95\textwidth]{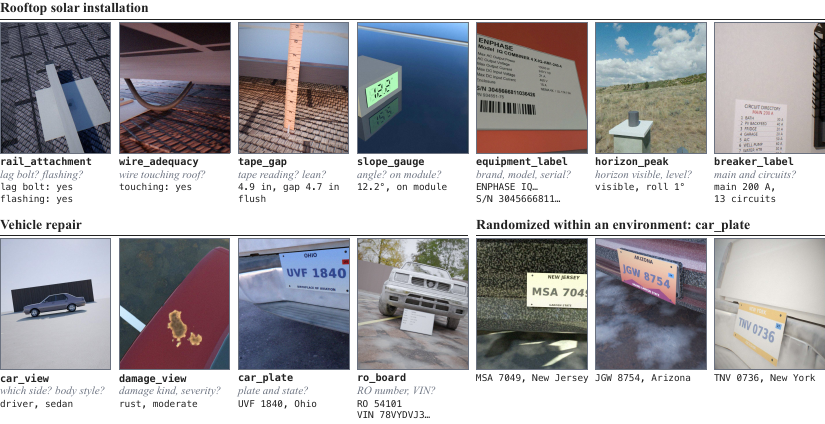}
\caption{The eleven environments of \bench{} with one scorer-acceptable frame from each. Because each scene is rendered in Blender, each frame can be scored programmatically. Under each frame: a short form of the environment's documentation question and the ground truth.}\label{fig:showcase}\end{figure}

It has been a long-term goal to create AI agents that augment human capabilities in the physical world, from early augmented-reality systems for manufacturing and maintenance~\citep{caudell1992augmented,feiner1993knowledge,henderson2011exploring} to recent egocentric assistants~\citep{wang2023holoassist,grauman2024egoexo4d,huang2025vinci}. These agents can help novice human workers perform tasks that require knowledge, e.g., showing them where to find a particular part of a machine, or reduce documentation time for expert workers by converting a recording of their task performance into a structured report. We expect these capabilities to be of increasing interest.

Towards this goal, we evaluate performance of current vision language models (VLMs) on two kinds of tasks (Figure~\ref{fig:showcase}): \emph{acceptability} and \emph{documentation}. Acceptability tasks ask, given a natural language specification, whether an image satisfies it, for the purposes of documenting the state of the environment (e.g., to show that work was performed). For example, in solar installation, an installer may request a sharp, close-up picture that shows the flashing and lag bolt for every L-foot on a solar module mounting rail (glossary: Appendix~\ref{app:glossary}). A verdict must intersect several conditions at once, some of them geometric (is the tape seated flush? is the gauge on the module?). Documentation tasks ask, given a natural language specification, to extract certain information from the environment that is needed to document the environment---for example, reading the height of a solar module above the roof off of a tape measure.

Evaluating the capability of agents on these tasks is challenging and expensive. In-field trials give the best ecological validity, but they are resource intensive. Archived in-field data exists in high volume, but suffers from two issues. First, it is restricted in access, limiting reproducibility and transparency. Second, it is imprecisely overseen: a single task may generate a huge volume of data (e.g., thousands of photos) with a human providing only a single label for an entire unit (\emph{packet}), under business pressure, and often assisted by VLMs. The weakness of VLMs at certain visual tasks makes VLM-based processing of archived field data problematic. This creates mechanistic uncertainty---it is unclear what a VLM \emph{should} be able to conclude from each image.

We therefore take a simulation approach. \bench{} renders eleven environments in Blender (Figure~\ref{fig:showcase}), each with an associated acceptability and documentation task, from rooftop solar installation (seven) and vehicle repair (four). This overcomes both issues---we release the rendered benchmark publicly under a permissive license and task success is programmatically evaluable using the environment state. This programmatic evaluation is challenging in the same way alignment to human preferences is---humans may have different views on how the natural language specification should be translated into requirements. Nonetheless, on held-out frames the scorer agrees with the majority of our human raters on 82\% of frames, close to the 76--92\% at which an individual rater agrees with the others (Section~\ref{sec:metrics}). We evaluate \nModels{} models on a benchmark of 1,378 frames from our simulator.

While simulation-based benchmarking has impressive flexibility, it is often difficult to know whether the results translate to the real world. We are able to estimate this gap through evaluations on in-field images procured from real-world physical work tasks, and real world reenactments (for one measuring task, which we release publicly). Model rankings transfer on all three matched documentation questions: across \corrTapeN{} models, Spearman $\rho \geq 0.63$ per question at $p \leq 0.002$. This is evidence that the simulated tasks predict which models do well on real photographs.

Our results provide insights into both what VLMs can perceive and how they judge. Documentation scores are high when the answer is text in frame (labels, plates, directory cards), except for long or unusual strings, and low on measurement (reading a tape at the crossing) and on judgments about small parts or relations (whether a tape leans, whether a lag bolt is present). On acceptability, the best models score about as well as a single non-author rater (F1 0.61 against \humanFone{}). We trace acceptability errors in three environments where the acceptability and documentation tasks share a field. There, models apply the requirement consistently to what they report seeing, but they misread the scene often. For example, they judge \misTapeRate\% of upright tape measures as tilted, and these misreadings account for most of their wrong rejections. For seven of the nine models we re-asked, the misreadings do not change when the verdict is asked last or not at all.

\paragraph{Contributions.}
\begin{enumerate}[leftmargin=*,itemsep=2pt]
\item A set of eleven environments from two kinds of real-world physical work, each with a human-validated programmatic acceptability scorer for a particular natural language task and a programmatic documentation scorer. Since the environments are simulators, not image sets, others can extend them to new tasks and to closed-loop guidance.
\item An evaluation of \nModels{} vision-language models on 1,378 frames from those environments, reporting acceptability against the scorer, documentation accuracy against the scene's ground truth, the cost of the whole set at list prices, and a comparison with a single human rater.
\item A sim-to-real comparison on three matched documentation questions using real photographs (a public tape-measure set that we release, and two proprietary sets of vehicle photographs), showing that rankings on rendered frames transfer to real photos.
\item An analysis of what makes the tasks hard for models, showing that in the three environments where we can trace them, most wrong rejections come from misreading the scene, not the natural language requirement.
\end{enumerate}

\section{Related Work}

\paragraph{Assistants and datasets for physical work.} Large egocentric corpora have the model observe the worker. Ego4D and Ego-Exo4D~\citep{grauman2022ego4d,grauman2024egoexo4d} and procedural datasets such as Assembly101, IndustReal and CaptainCook4D~\citep{sener2022assembly101,schoonbeek2024industreal,peddi2024captaincook4d} label steps, mistakes~\citep{lee2024egoper} and proficiency. HoloAssist~\citep{wang2023holoassist} records a remote instructor guiding a worker in real time, and portable egocentric VLM assistants are beginning to appear~\citep{huang2025vinci}. For inspections, VLMs have been evaluated as construction-safety inspectors~\citep{chen2025constructionsite10k} and as unified industrial-inspection models, where they proved brittle against task-specific baselines~\citep{wallace2025inspectvlm}. These datasets are video-centric and label the work---they don't ask whether a captured view is adequate proof that the work was done. We found no public dataset of worker-submitted proof-of-work photos judged for acceptability.

\paragraph{Synthetic environments and sim-to-real.} Rendered data with exact ground truth has a long history as training data~\citep{richter2016playing,roberts2021hypersim} and as a diagnostic test bed, e.g.,  CLEVR~\citep{johnson2017clevr} and Super-CLEVR~\citep{li2023superclevr}, which varies scene factors independently. Generators include Kubric, BlenderProc and Infinigen~\citep{greff2022kubric,denninger2023blenderproc2,raistrick2023infinigen}, and embodied simulators include ProcTHOR and Habitat~\citep{deitke2022procthor,savva2019habitat}. Domain randomization~\citep{tobin2017domainrandomization} and synthetic spatial training that improves real-image benchmarks~\citep{ray2025sat} show that variation can improve sim-to-real transfer. We know of no prior work that measures whether VLM \emph{rankings} on a rendered benchmark match rankings on real photos of the same task.

\paragraph{Judging image quality.} No-reference image quality assessment scores perceptual quality from statistics~\citep{mittal2012brisque,mittal2013niqe} or learned models~\citep{ke2021musiq}, and recent work puts the question to VLMs directly: CLIP-IQA~\citep{wang2023clipiqa} with antonym prompts, Q-Bench~\citep{wu2024qbench} as a low-level-vision benchmark, and Q-Align and DepictQA~\citep{wu2024qalign,you2024depictqa} for calibrated scores and language explanations. These judge quality independent of content. Task-defined acceptability appears in mobile document capture, where a frame is good if downstream OCR succeeds~\citep{burie2015smartdoc,li2018cgdiqa,bulatov2020midv2019}, and in insurance imaging, where customer photos feed damage estimation~\citep{mallios2023vehicledamage} and VLMs are benchmarked on claims imagery~\citep{lin2025insmmbench}, but there the photos are assumed adequate.

\paragraph{Scoring an image against a specification.} TIFA and Davidsonian Scene Graph~\citep{hu2023tifa,cho2024dsg} decompose a prompt into questions answered by a VQA model, VQAScore~\citep{lin2024vqascore} asks a single yes/no, and GenEval~\citep{ghosh2023geneval} runs a hand-written detector pipeline over object state, reporting agreement with annotators close to inter-annotator agreement. Visual programming~\citep{gupta2023visprog,suris2023vipergpt,cho2023vpeval} and LLM-written reward code over simulator state~\citep{ma2024eureka} generate the verifier itself. Our scorers are hand-written programs over privileged scene state, closest in spirit to GenEval, and we validate them against human ratings as that literature does. We do not use a VLM as the reference scorer, partly because studies of multimodal LLMs as judges find absolute scoring inconsistent even for frontier models~\citep{chen2024mllmjudge}.

\paragraph{Fine-grained perception in VLMs.} BLINK~\citep{fu2024blink} finds frontier models near chance on tasks humans solve at a glance, MMVP~\citep{tong2024eyes} traces failures on orientation, counting and small details to the CLIP encoder, and HallusionBench~\citep{guan2024hallusionbench} separates answering from priors from misreading the image. Spatial benchmarks report that metric questions are among the hardest~\citep{yang2025thinking,chen2024spatialvlm}, and PhysBench~\citep{chow2025physbench} covers physical properties and relations broadly. Reading text is well studied, e.g., through scene-text and document VQA~\citep{singh2019towards,mathew2021docvqa} and OCRBench~\citep{liu2024ocrbench,fu2024ocrbenchv2}, which singles out non-semantic strings such as serial numbers as a weakness. Closest to our tape and gauge tasks, MeasureBench~\citep{lin2026measurebench} benchmarks reading values off instruments in real and synthesized images and finds that frontier VLMs struggle, while specialist readers trained on synthetic gauges remain competitive~\citep{leonalcazar2024gauges,mueez2026gauge}.

\section{Environments}
\label{sec:envs}
\label{sec:bench}

We construct 11 environments in Blender 4.5.12 in two settings: rooftop solar installation (seven) and vehicle repair (four). Each environment uses a shared model of the phone camera: a phone is drawn per scene (main lens 24--28\,mm full-frame equivalent, sensor width 5.6--9.8\,mm, f/1.6--2.4, and on some phones ultra-wide and telephoto lenses and digital zoom), each frame is rendered in landscape or portrait at that phone's effective focal length, and the aperture is set from the physical lens so that Cycles renders a phone's depth of field. 70\% of frames are focused on the target with a steady hand, 15\% have the autofocus locked elsewhere and 15\% have hand shake over a shutter time drawn log-uniformly from 1/500 to 1/25\,s. Sensor noise, white balance, tone mapping, chromatic aberration, vignetting and a JPEG re-encode are applied after the render.

Each environment has an acceptability and documentation task. The acceptability tasks ask whether a photo satisfies a natural language condition. The documentation tasks ask the model to write some property of the environment to a structured JSON format. Example prompts are in Appendix~\ref{app:prompts}.

\paragraph{Why these tasks?} First, they are derived from tasks where VLMs fell short on real field images. Second, they test models along many axes. For example, \env{rail\_attachment} asks whether a close-up shows the flashing and lag bolt of a single roof attachment, testing recognition of small hardware in clutter. Other tasks ask the model to read text and capture an image that shows that text clearly to a human. Finally, while there is a body of VLM work in structured settings such as kitchens and warehouses that have predictable lighting, geometry, and objects, field-facing tasks such as rooftop solar installation and vehicle repair are unconstrained and high variance, and not rigorously studied.  Together, these tasks allow us to assess perception and understanding in a nuanced way.

\subsection{Rooftop Solar Installation}
Six of the seven solar environments are on the tilted roof of a house where a solar array is being or has been installed. Five share one procedural roof scene: the roof rectangle and pitch, the shingle courses, the house body under the roof (lap siding or brick), a gable or shed roof, a chimney (about half of the scenes), a ladder leaning on the eave, the array layout and the sun are all drawn per scene from the generator's distributions. The background is a sky-only HDRI panorama with the sun lamp aligned to the panorama's own sun. \env{horizon\_peak} uses one of 21 CC0 house meshes instead, re-covered with the same procedural shingle courses and trims, placed inside a full $360^\circ$ panorama. The seventh, \env{breaker\_label}, is set at the house's load center on a wall.

We follow an installation in order. The array's rails are held to the roof by L-feet, each lag-bolted into a rafter through a sheet of flashing slipped under the shingles which, together, anchor the rail and prevent leakage. \env{rail\_attachment} asks for a close-up of one attachment that shows the L-foot, its flashing and the lag bolt, and asks whether the bolt and the flashing are present. Under the modules, the wire should not touch the shingles: \env{wire\_adequacy} asks for a photograph of the wire's lowest point, and asks whether any wire touches the roof (the requirement also mentions hard bends, which the scorer does not check separately). The module must stand clear of the roof: \env{tape\_gap} asks for a tape measure standing up to the module frame, and asks for the number at the crossing, the true perpendicular gap and whether the tape leans. The module's slope is recorded with a digital angle gauge: \env{slope\_gauge} asks for the gauge sitting on the module, aligned with the slope, with the module visible (the scorer also requires some of the module's edge in view, since alignment cannot be judged without it), and asks for the reading. Each piece of equipment is photographed by its nameplate: \env{equipment\_label} asks for a legible nameplate and for its brand, model and serial number and what it is on. From the ridge, the installer photographs the skyline in every direction, which records the obstructions that shade the array. \env{horizon\_peak} asks for a photo, taken from the peak, of the skyline in one direction with a little of the house's own roof along the bottom, and asks whether the horizon is visible and level. Finally, the array ties into the house's electrical panel: \env{breaker\_label} asks for a square-on, legible close-up of the circuit directory card and for its full contents, the main breaker amperage and every circuit's name and amperage.

\subsection{Vehicle Repair}
The photographer is in a parking lot with multiple vehicles from a set of nine meshes, with the subject car identified by color and body style
(``the silver sedan''). A damaged vehicle arriving at a repair shop is first documented as a whole. \env{car\_view} asks for a particular face of the car (front, rear, driver or passenger side) straight on, with the whole car in frame from about two car lengths, and asks which face is shown and the body style. The damage is then photographed up close: in \env{damage\_view}, a procedurally generated dent, scratch or rust mark on a panel must be photographed square-on from about an arm's length, with part of the adjacent panel in frame for context, and the model is asked the kind of damage and its severity (minor, moderate or severe, by surface area). The vehicle's identity is recorded: \env{car\_plate} asks for the mounted license plate close up and legible, and for its characters and the state named on it. Finally the shop issues a repair order, written with the vehicle's VIN on a hand-held board and photographed with the car so the two are tied together: \env{ro\_board} asks for the board, legible, with the car in the same frame, and for the RO number and the VIN.

\section{Building the Benchmark Set and Acceptability Scorer}
\label{sec:metrics}

For each of the 11 scenes, we render 100 frames with the camera drawn from that environment's pose distribution. We perform additional positive sampling for the seven environments where random frames are rarely acceptable, for a total of 1,450 initial frames. In \env{ro\_board}, 29 frames were withdrawn because of a defect in the board texture, leaving 1,421 frames.

We define an acceptability scorer for each environment, which is an intersection of acceptability terms.
Some acceptability terms are computed directly from the scene after the camera is placed, including in-frame and occlusion by ray casting, distances, view angles, and pixels per unit length. Another set of terms is read off the rendered frame itself, including sharpness at the target, the text's contrast against its paper and its blown-out fraction, the wire's pixel count and edge contrast, and the label's paper luminance. Acceptability is orthogonal to work completeness---unacceptable work can be acceptably documented. Documentation answers are scored against the scene's ground truth. Details are in  Tables~\ref{tab:envs} and~\ref{tab:terms}. We call frames that pass the acceptability scorer \emph{scorer-acceptable}.

We tune the scorer to simulate human judgment. Six raters (four authors and two colleagues) judged 30 random frames per environment for acceptability, against the requirement text the models see and without seeing the scorer's verdict: two raters judged all of them, three judged 20--25 per environment, and one author judged only \env{tape\_gap} and \env{damage\_view} (4.0--5.1 ratings per frame, Table~\ref{tab:human_ratings}).\footnote{Ten of the rated \env{ro\_board} frames were among those later withdrawn; their ratings are kept in the scorer tuning and in the agreement figures below.}
We split this set once, by frame, into 60\% used to adjust the scorer and set the gray zones (described below) and 40\% looked at once, after the scorer was frozen. On each held-out frame with at least three ratings we take the raters' majority as the reference (95 frames after excluding ties). The tuned scorer agrees with that majority on 82\% of frames (95\% CI 75--90\%). It errs on the strict side: it accepts 4 of the 62 frames the majority rejects and rejects 13 of the 33 that the majority accepts. An individual rater agrees with the majority of the \emph{other} raters on 76--92\% (67\% for the author who rated two environments, on 18 frames, Table~\ref{tab:human_ratings}). We opted to drop the acceptability task for \env{car\_plate}, where human raters disagreed on small but legible plates, and the acceptability task for  frames of \env{rail\_attachment} that clearly show an attachment with its lag bolt or flashing missing, due to task ambiguity. We removed the scorer requirement that the \env{horizon\_peak} horizon be level, which the requirement text did not ask for. After these reductions and the gray zones below, 1,378 frames remained. Of these, 1,265 are scored for acceptability, and the 547 scorer-acceptable frames (including those of \env{car\_plate}) are scored for documentation.\footnote{Tape leaning is scored on all 143 \env{tape\_gap} frames as lean can be judged from all frames, and all scorer-acceptable frames have no lean.}

\paragraph{Acceptability gray zones.} Natural language descriptions that are used in the field are ambiguous (what does ``arm's length'' mean?). We want to avoid marking down models that reasonably interpret natural language. Where the wording leaves a range open, or our human raters disagree with each other over it, we mark that range of the term as gray (four zones in Table~\ref{tab:envs}). For example, in \env{damage\_view}, which asks for a photo from about arm's length, camera distances between 1.5 and 3\,m are gray. A frame whose verdict depends only on a gray term is not scored for any model. We drop 43 frames from the acceptability set this way.

\paragraph{Other metrics.} We report the cost of the whole set at the developer's own list API price (Table~\ref{tab:static_v2_prices}), with thinking tokens priced at the output rate.\footnote{Cache benefits would be small due to the input tokens being dominated by the image. We report numbers without cache benefits.} We served three models ourselves (Gemma 4 E4B, LFM2.5-VL-3B, Qwen3.5-4B), which we do not attempt to price due to differences between self-hosting and API costs. We report the mean number of thinking tokens per item (where a provider reports only an output count, the count minus the tokens of the reply we received) and the fraction of answers that match the stated grammar.

\section{Real-World Data}
\label{sec:real}

For three of the tasks we have approximate real-world equivalents, which let us estimate the sim-to-real gap.

\paragraph{Reading tape measures.} We find that reading the measured distance off a tape measure precisely is one of the most challenging synthetic tasks. To test whether this result transfers, we photographed a tape measure held against everyday objects (monitors, a laptop, benches, tables, a fence, a whiteboard, a car battery) and recorded the ground truth for each measurand with a caliper or a careful human tape read: 133 readings on 84 photos of 35 items across 12 scenes, which we release. Each reading is asked with a prompt of the same form as \env{tape\_gap}'s, naming the object and the measurement in place of the module and the gap, and scored as read within $\nicefrac{1}{8}$\,in of the ground truth.

\paragraph{Assessing car damage.} We use photographs of damaged vehicles from the real world. For \env{car\_view}'s face question, one author labeled 117 real frames with every side and third of the car they show---the model's answer was marked correct if it names one of them. This is similar to \env{car\_view}'s documentation task. For \env{damage\_view}'s severity question, one author rated the 63 photos of damaged panels on the documentation task's severity scale.

\section{Experiments}
\label{sec:exp}

We evaluate the models shown in Table~\ref{tab:static_v2_totals}, including: 23 through their providers' APIs under provider default settings, and three small open-weight models (Gemma 4 E4B~\citep{gemma4}, LFM2.5-VL-3B, Qwen3.5-4B) that we served ourselves in bf16 at the model-card recommended settings. We evaluate Gemma 4 E4B with thinking on and Qwen3.5-4B in its non-thinking mode: in thinking mode Qwen3.5-4B used a mean of 3,064 thinking tokens per item on a 180-item sample, about four times Gemma 4 E4B (740). We report the acceptability F1 and documentation score vs.\ cost to complete the entire set in Fig.~\ref{fig:static_cost}. More experimental details, model performance statistics, and confidence intervals are in Appendix~\ref{app:static}. Four models were not run on the real photographs due to logistical issues: Qwen 3.8 Max, Muse Spark 1.3, Kimi K3~\citep{kimik3}, and DeepSeek V4.1 Flash~\citep{deepseekv41flash}.

\begin{figure}[t]\centering
\begin{minipage}[c]{0.375\textwidth}\centering\includegraphics[width=\linewidth]{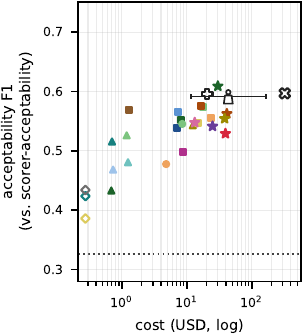}\end{minipage}\hfill
\begin{minipage}[c]{0.375\textwidth}\centering\includegraphics[width=\linewidth]{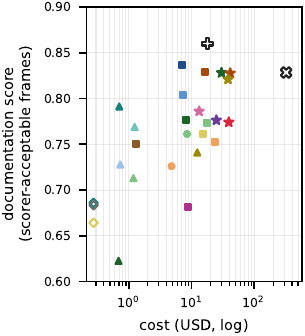}\end{minipage}\hfill
\begin{minipage}[c]{0.22\textwidth}\centering\includegraphics[width=\linewidth]{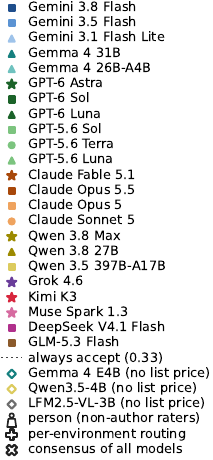}\end{minipage}\\[4pt]
{\centering\includegraphics[width=0.75\textwidth]{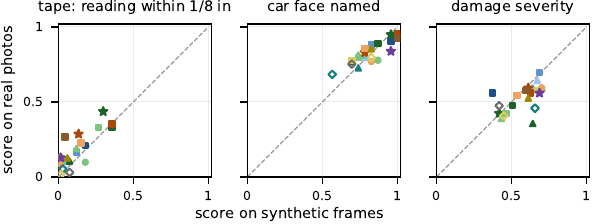}\par}
\caption{\textbf{Top:} acceptability F1 against scorer-acceptability (left) and documentation score on scorer-acceptable frames (right) against the cost of the whole set. Color marks the developer and shade the generation (darker is newer; Gemma 4 26B-A4B is a lighter shade only to tell it from Gemma 4 31B) and shape is a rough size tier ($\bigstar$ flagship, $\blacksquare$ large, $\bullet$ mid, $\blacktriangle$ small, $\blacklozenge$ 3--4B). The dotted line is the ``always accept'' baseline; the hollow markers at the left edge are the small open-weight models we served ourselves. \raisebox{-0.25ex}{\includegraphics[height=1.6ex]{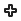}} routes each environment to the model that scored best on it (chosen on the other half of the frames) and \raisebox{-0.25ex}{\includegraphics[height=1.6ex]{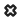}} takes the majority verdict and answer of all models. \raisebox{-0.25ex}{\includegraphics[height=1.6ex]{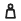}} marks a single non-author rater's F1 (95\% CI \humanFoneLo--\humanFoneHi) at the cost of one human pass over the set (\$\humanCostLo--\humanCostHi). \textbf{Bottom:} documentation score on synthetic frames (scorer-acceptable; for severity, frames where the damage is readable) against real photographs on the three matched questions.}
\label{fig:static_cost}
\end{figure}

\subsection{Results}

On synthetic data, models spread widely on both tracks (acceptability F1 0.39--0.61, documentation 0.62--0.84), with the two rankings in approximate agreement (Spearman $\rho = \rhoFoneDoc$, $p \pFoneDoc$). On each track, we call the leader and the models whose paired 95\% interval against it includes zero the \emph{top group} (Table~\ref{tab:static_v2_paired}). On acceptability, GPT-6 Astra leads (0.61, \$30.05), with Claude Opus 5.5 (0.58, \$16.36), GLM-5.3 Flash (0.57, \$1.30) and Gemini 3.5 Flash (0.56, \$7.36) in the top group. On documentation, Gemini 3.8 Flash leads (0.84, \$7.16), with Claude Opus 5.5 (0.83, \$16.36), GPT-6 Astra (0.83, \$30.05), Claude Fable 5.1 (0.83, \$41.35) and Qwen 3.8 Max (0.82, \$38.19) in the top group. GPT-6 Astra and Claude Opus 5.5 are the only models in both top groups. The best open-weight model on acceptability is GLM-5.3 Flash, and on documentation the much smaller Gemma 4 31B (0.79, \$0.70). Routing across models helps documentation but not acceptability: routing each environment to the model that scored best on it (chosen on the other half of the frames) reaches a documentation score of 0.86 (\$17.90), above every single model, but an F1 of only 0.60 (\$20.60). The majority of all models reaches F1 0.60 and documentation 0.83 at \$322---on par with or worse than the best single model.

On the cost axis, the models fall into three separated groups: those costing over \$4 for the set (\$4.80--41.35), six costing under \$2 (\$0.69--1.30, which we call moderately priced), and the unpriced models we served ourselves.
The moderately priced group contains three open-weight models (Gemma 4 31B, Gemma 4 26B-A4B and GLM-5.3 Flash) and three closed-weight models (GPT-6 Luna, Gemini 3.1 Flash Lite and GPT-5.6 Luna). The top group on acceptability within the moderately priced band is GLM-5.3 Flash (0.57, \$1.30) and GPT-5.6 Luna (0.53, \$1.18), and on documentation, Gemma 4 31B (0.79, \$0.70) alone. Two of these three top group models are open-weight and thus make especially attractive targets for fine-tuning.

The unpriced band is on par with the lowest scoring priced models on both tracks, with LFM2.5-VL-3B and Gemma 4 E4B at the top of it (F1 0.43 and 0.42, documentation 0.68 and 0.69). These models are exciting because they can run on-device, with low latency and no network dependence.

\paragraph{Real-world images.} Figure~\ref{fig:static_cost} (bottom) and Table~\ref{tab:real_vs_sim} give, per model, the synthetic and real scores on the three matched questions. We see the tasks have broadly consistent difficulties between real and synthetic environments, with tape reading hardest on both, followed by damage severity, followed by naming the seen face of the car (which is saturated for top models). The rankings transfer, with Spearman $\rho$ of 0.63--0.79 ($p\leq 0.002$) with the worst transfer on damage severity.\footnote{On \env{damage\_view}, the real frames are not solely scorer-acceptable---thus, we use a sharpness-and-visibility-only condition to create the synthetic comparison set.}

\paragraph{Human raters.} Using only the raters who were not involved in designing the scorer, a single rater scores an F1 of \humanFone{} (95\% CI \humanFoneLo--\humanFoneHi) against the scorer. Individual raters are noisy---the scorer tracks their majority. We cannot distinguish the strongest models from a single human rater on acceptability. We perform an approximate cost comparison. Our raters took 8--12\,s per frame (including the documentation fields) while doing other work; they are not expert raters and did not know they were being timed, so we allow for faster work. At 3--12\,s per item and \$\humanWageLo/h (a somewhat arbitrary lower bound) to \$\humanWage/h~\citep{bls2026oews}, one human pass over the \humanItems{} items costs roughly \$\humanCostLo--\humanCostHi, against \$16.36 for Claude Opus 5.5 and \$30.05 for GPT-6 Astra. Across generations, models improve on score and cost at once, so if the trend holds they will overtake a single rater on these tasks. Their remaining weaknesses are specific, and we turn to them next.

\section{Analysis}
\label{sec:hard}

\begin{figure}[t]\centering\includegraphics[width=0.9\linewidth]{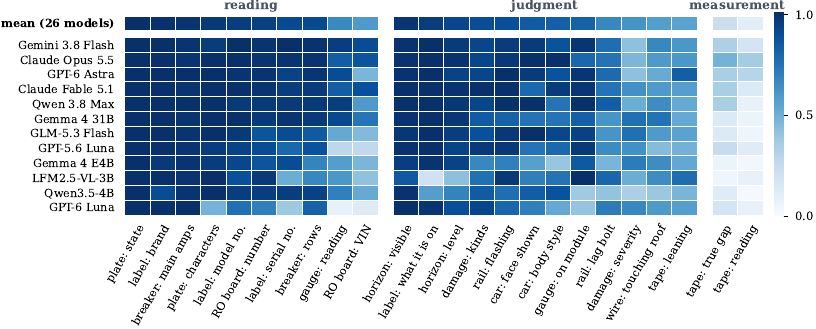}
\caption{Documentation score by field (columns) and model (rows) on scorer-acceptable frames (tape leaning: all \env{tape\_gap} frames). Columns are grouped by skill and sorted by the mean over all \nModels{} models (top row). Rows are the models on a cost frontier of Figure~\ref{fig:static_cost}, the other models in the documentation top group and the three models we served, sorted by documentation score.}\label{fig:field_scores}\end{figure}

\paragraph{What is easy and hard for models?} Figure~\ref{fig:field_scores} divides the \nFields{} documentation fields into three groups: \emph{reading} (transcribe text or digits), \emph{judgment} (a categorical scene state) and \emph{measurement} (for reading tape measures) (details: Table~\ref{tab:field_scores_full}). Models are very strong at reading: eight of the ten reading fields score 0.90--1.00 averaged over models. Long or unusual strings are still hard on average: the 17-character VIN (mean 0.59) and the gauge's seven-segment display (0.66), though Gemini 3.8 Flash scores 0.89 and 0.95. On judgment fields, those whose answer fills the frame are easy: what the label is on (mean 0.95), whether the horizon is visible (0.98) or level (0.91), the kind of damage (0.89), the car's face (0.83) and body style (0.81). The hard ones concern a small part or a relation: whether the lag bolt is present (0.67, best 0.82), how severe the damage is (0.61, best 0.75), whether the wire touches the roof (0.57, best 0.74) and whether the tape is leaning (0.55, best 0.82). Reading a tape measure is the hardest tested task: the number at the crossing is within $\nicefrac{1}{8}$\,in 11\% of the time on average (best 36\%, Opus 5.5 and GPT-6 Sol), and the true gap within $\nicefrac{1}{4}$\,in 23\% (best 48\%, Opus 5.5); as we saw, this difficulty carries over to real photographs. Across models, reading score is not detectably correlated with judgment (Spearman $\rho = \rhoReadingJudgment$, $p \pReadingJudgment$), while judgment and measurement are strongly correlated ($\rho = \rhoJudgmentMeasurement$). Acceptability F1 follows judgment ($\rho = \rhoFoneJudgment$) and measurement ($\rho = \rhoFoneMeasurement$), not reading ($\rho = \rhoFoneReading$, $p \pFoneReading$).

\paragraph{Drivers of difficulty within an environment.} In \env{tape\_gap} the models read the tape's angle in the image rather than against the module. The tape is within $5^\circ$ of upright on every scorer-acceptable frame, yet the models call it tilted \tapeSaidTilted\% of the time, and the rate rises with the roll of the camera. The rise appears in \tapeRollRisesModels{} of the \nModels{} models and in \tapeRollRisesFrontier{} of the \nFrontier{} on the cost frontiers. Close shots with little roof for reference are the worst (Spearman $-0.59$ with roll, $-0.51$ with the apparent blade size). Reading fields are governed by resolution and viewing angle, as expected: the VIN on the RO board tracks pixels per text row ($+0.71$), the RO number the yaw of the board ($-0.51$), the breaker directory the viewing angle ($-0.44$). Two judgment fields are governed by visibility and vantage: whether the lag bolt is present tracks how much of the mount is unoccluded ($+0.56$), and whether the gauge sits on the module tracks camera pitch ($+0.51$).

\paragraph{Do models misread the requirement or the scene?} Acceptability F1 follows the geometric judgment fields, not reading. We analyze this through three environments where the acceptability and documentation tasks overlap: whether the tape is tilted, whether the gauge is on the module, whether the lag bolt and flashing are present. On scorer-acceptable frames the property always holds, so a model that reports the defect has misread the scene. Models apply the requirement to what they read---they reject \misRejMin--\misRejMax\% of the frames on which they report the defect---but they misread often, and the misreadings come with most of their wrong rejections: \misTapeShare\% in \env{tape\_gap}, \misSlopeShare\% in \env{slope\_gauge} and \misRailShare\% in \env{rail\_attachment} (Table~\ref{tab:accept_by_answer}).

\paragraph{Is the misreading real, or a justification of the verdict?} The reply gives the verdict first, so a model could reject a frame for another reason and then report a defect to justify it. We re-asked 150 frames from these environments to nine models with the verdict last and with no verdict at all (Appendix~\ref{app:order}). For seven, the defect reports do not change beyond run-to-run noise. The two cheapest, Gemini 3.1 Flash Lite and Gemma 4 31B, report fewer defects without the verdict (Flash Lite about half as many). For them, asking for the verdict first worsens their documentation answers.

\paragraph{Documentation on scorer-rejected frames.} On the random frames, the documentation score is \docEngineAdq{} on scorer-acceptable frames and \docEngineInadq{} on the rest, and every model scores higher on scorer-acceptable frames (by \docEngineDeltaMin{}--\docEngineDeltaMax{}). The gap is largest on the reading tasks (+0.58 to +0.70) and smallest in \env{horizon\_peak} (+0.09), \env{car\_view} (+0.07), \env{wire\_adequacy} (+0.01) and \env{tape\_gap} ($-0.17$). In \env{tape\_gap} the tape leans on \tapeLeanShareInadq\% of scorer-rejected frames, where the models' tendency to call the tape tilted is usually right.

\section{Conclusion}

We introduced \bench{}, eleven rendered environments from rooftop solar installation and vehicle repair. The environments are simulators, not image sets, so they extend to new tasks and to closed-loop guidance. Each environment poses two tasks a field agent needs, both scored programmatically from the scene: judging whether a photograph is adequate proof of work and reading the task-relevant state from it. Our observations show that reading text in frame is largely solved. The weaknesses are in reading measurements and in judging small parts and spatial relations. Acceptability, however, is harder to pin down, since raters disagree with each other, and the best models are level with a single non-author rater. In the three environments where we can trace acceptability errors, most wrong rejections come from misreading the scene. Rankings on rendered frames transfer to real photographs on all three matched documentation questions.
\subsection*{AI use statement}

In this work, we used generative AI tools for the following tasks with required disclosure: proposing or refining hypotheses; designing or providing feedback on research methodology or experiments; implementing methods; cleaning and reformatting data; and interpreting results. We have not used generative AI tools to generate synthetic data; assist with translation; or support qualitative and thematic data analysis. Helping develop theoretical models or conceptual frameworks, formulating mathematical claims, providing critical ingredients for proving mathematical claims, assisting in the writing of proofs are not applicable to this work.
Additionally, we used generative AI tools for tasks with recommended disclosure: creating and editing software code, creating or modifying scientific figures or images, drafting parts of a research paper, identifying and summarizing related literature, and editing a research paper to improve readability. We have reviewed all AI-assisted work. Every scorer threshold was set against human ratings collected without AI involvement, the model runs and their analysis scripts were checked by the authors, and AI-proposed interpretations were kept only where the authors could verify them. We take responsibility for the final content of this work, including text, claims or artifacts produced with the aid of generative AI.

\subsection*{Ethics statement}

\paragraph{Human ratings.} The acceptability ratings used to validate the scorers and to compare models against people were produced by four of the authors and two colleagues, rating rendered frames of inanimate scenes; one author also labeled the vehicle photographs of Section~\ref{sec:real} with the face shown and the damage severity. No personal data was collected from the raters, so no ethics review was sought.

\paragraph{Real photographs.} The tape-measure photographs were taken by an author of their own objects; we release them. The vehicle damage photographs are used with permission. We use them only to score model answers, do not release them, do not reproduce them in the paper, and sent them only to providers whose data-retention terms met ours.

\paragraph{Conflicts of interest.} The benchmark measures the general capability of third-party models rather than any product produced by the authors, and the API models were run through their public APIs at list prices and the three open-weight models on rented GPUs.

\paragraph{Assets and licenses.} Every third-party 3D asset in the environments is CC0 or CC-BY and is listed with its license and attribution in the released dataset's attribution file.

\paragraph{Potential harms.} A system that judges photographs of completed work could be used to pressure workers or to automate rejection of legitimate work. We argue that better evaluation of model capabilities is important for making societal decisions about the use of AI in human work.

\bibliography{refs}
\bibliographystyle{iclr2027_conference}

\appendix
\begin{table}[p]
\centering\scriptsize
\caption{Environment, task and scorer summaries. Every capture task additionally requires \emph{sharp at target} (defocus and motion blur at the target $\le 5$\,px; $6$\,px for \env{car\_plate}, \env{car\_view}, \env{ro\_board}, \env{damage\_view}). \emph{Acceptable} is the fraction of the random field-pose frames of each environment that the scorer accepts. Documentation question is a summary of the task---not the prompt given to the model.}
\label{tab:envs}
\begin{tabularx}{\textwidth}{@{}l l >{\hsize=.75\hsize}X >{\hsize=1.25\hsize}X r@{}}
\toprule
Environment & Scene & Documentation question & Acceptability terms & Acceptable \\
\midrule
\env{rail\_attachment} & roof & At this roof attachment, is a lag bolt present on the L-foot, and is flashing installed under it? & attachment in frame ($\ge$ 90\% of it); attachment unoccluded; bolt location unoccluded ($\ge$ 50\% of the spot where the lag bolt sits, or would sit, visible); fills the frame (bbox $\ge$ 2\%); not a wide shot ($\le$ 1.8 m); \emph{not scored when the lag bolt or flashing is missing} & 0.12 \\
\env{wire\_adequacy} & roof & Is any wire or cable touching the roof surface? & a lead wire in frame ($\ge$ 50\% of it); that wire's lowest point visible; that wire unoccluded ($\ge$ 30\% of it); wire visible to the eye: wire pixels $\ge$ 0.05\% of the frame; wire stands out: edge contrast $\ge$ 0.03; close enough to judge contact (cam within 2.5 m of that wire's low point) & 0.39 \\
\env{tape\_gap} & roof & What number is on the tape where the module frame crosses it, what is your estimate of the true perpendicular module-to-roof gap in inches, and is the tape leaning rather than standing flush? & crossing in frame; crossing unoccluded; close enough to read 1/16in ticks ($\le$0.65 m); tape seated flush (tilt $\le$ 5$^\circ$); photographed from the front of the blade; \emph{gray: a lean over 5$^\circ$ that shows as at most 5$^\circ$ in the image} & 0.17 \\
\env{slope\_gauge} & roof & What does the angle gauge read, in degrees, and is it sitting on the module (glass or frame) or on bare roof? & display in frame; display unoccluded; digits legible: $\ge$ 14 px digit height; display readable angle (view angle $\le$ 65$^\circ$); gauge on the module, not bare roof; display not blown out ($\le$ 30\% saturated); gauge aligned with the slope (|yaw| $\le$ 15$^\circ$, else it under-reads); array context: module glass visible in frame; module edge in view to judge alignment ($\ge$ 300 px of frame edge visible); text stands out in the render (Weber contrast $\ge$ 0.2); \emph{gray: |yaw| 15--45$^\circ$} & 0.25 \\
\env{equipment\_label} & roof & What does it say on the equipment label (brand, model number and serial number) and what kind of equipment is it on? & label fully in frame; label unoccluded ($\ge$ 95\% of it); legible: $\ge$ 10 px per text row; barcode resolvable: $\ge$ 1 px per module; square-on enough to OCR (view angle $\le$ 60$^\circ$); label not blown out ($\le$ 30\% saturated); site context: some of the equipment/mounting in view; label lit: paper luminance $\ge$ 0.05; text stands out in the render (Weber contrast $\ge$ 0.3) & 0.06 \\
\env{horizon\_peak} & roof & Is the horizon visible in this frame, and is it roughly level? & horizon in frame, in the middle 80\% of the image; taken from the peak: the house's own roof hides $\le$ 10\% of the horizon arc; skyline inside the frame: sky above it and the surroundings below it across $\ge$ 65\% of the width; level (roll $\le$ 10$^\circ$); own roof visible along the bottom of the frame ($\ge$ 4\% of the bottom quarter): proof it was taken from the peak; phone at or above the ridge line (camera height above the ridge $\ge$ 0 m) & 0.19 \\
\env{breaker\_label} & wall & What is the main breaker amperage and every circuit's name and amperage? & card in frame; card unoccluded; legible: $\ge$ 11 px per text row; square-on (view angle $\le$ 45$^\circ$); tight: card bbox $\ge$ 4\% of frame; text stands out in the render (Weber contrast $\ge$ 0.225); card not blown out ($\le$ 30\% saturated) & 0.05 \\
\env{car\_view} & car & Which side of the vehicle is this (front, rear, driver side, passenger side)? What is the vehicle's body style (sedan, SUV, etc.)? & whole car in frame; not three-quarter: |angle off the nearest face| $\le$ 30$^\circ$; about two car lengths (1.5–3 lengths); car fills $\ge$ 3\% of frame; \emph{gray: 3--4 car lengths} & 0.23 \\
\env{damage\_view} & car & What kind of damage (dent, scratch, rust) is this, and how severe is it? & damage in frame; damage unoccluded; square onto the panel (view angle $\le$ 45$^\circ$); an arm away (0.3–3.0 m); adjacent panel context ($\ge$ 10\% of a neighboring panel in frame); resolved: $\ge$ 3 px per cm at the damage; damage visible ($\ge$ 60 contrasting px in frame); \emph{gray: distance 1.5--3 m} & 0.38 \\
\env{car\_plate} & car & What is the license plate number and the state named on it? & plate in frame; plate unoccluded; legible: $\ge$ 12 px per character; straight-on (view angle $\le$ 45$^\circ$); fills most of the frame (bbox $\ge$ 2\%); text stands out in the render (Weber contrast $\ge$ 0.1); plate not blown out ($\le$ 30\% saturated); \emph{acceptability not scored} & -- \\
\env{ro\_board} & car & What is written on the repair order board? What is the body style of the vehicle (sedan, SUV, etc.)? & board fully in frame; board unoccluded; legible: $\ge$ 9 px per text row; square-on (view angle $\le$ 40$^\circ$); vehicle in the same frame ($\ge$ 50\% of the car in frame); board not blown out ($\le$ 30\% saturated); text stands out in the render (Weber contrast $\ge$ 0.3) & 0.18 \\
\bottomrule
\end{tabularx}
\end{table}

\section{Limitations}
We evaluate single frames, whereas workers would naturally produce a video. The scorer is one reading of each requirement, validated against six raters on 30 frames per environment, and it errs on the strict side. The human comparison uses non-expert raters, and its cost is a rough range. The error tracing covers wrong rejections in the three environments where acceptability and documentation share a field. The real-world comparison covers three documentation questions and no acceptability task. The eleven environments come from two kinds of work. We test general-purpose models zero-shot at provider defaults, not specialist readers or detectors, and model versions and prices change quickly.

\section{Glossary}
\label{app:glossary}
\begin{description}[leftmargin=*,itemsep=1pt]
\item[module] a solar panel. It is clamped to a pair of rails, and each rail is held to the roof by L-feet lag-bolted through flashing into the rafters.
\item[flashing] the metal sheet slipped under the shingles at each roof attachment so the lag-bolt hole does not leak; only its lower part is exposed.
\item[L-foot / lag] the bracket that holds the rail to the roof, and the lag bolt through it into the rafter.
\item[horizon from the peak] photos taken from the ridge in every compass direction so shading obstructions can be read from the skyline.
\item[apparent reading] the number on the tape where the camera sees the module edge cross it; differs from the geometric crossing by parallax.
\item[repair order board] a board on which the shop writes the repair order (RO) number and the vehicle's VIN, photographed together with the vehicle so the two are tied to each other.
\end{description}

\begin{table}[t]
\centering\small
\caption{Scoring primitives. The first group is computed in the scene after the camera is placed; the second reads the rendered frame (the target's pixels are located by an object-index mask rendered from the same camera).}
\label{tab:terms}
\begin{tabular}{@{}l p{9.4cm}@{}}
\toprule
Primitive & Definition \\
\midrule
in-frame / unoccluded fraction & Sample points on the target's surface (vertices \emph{and} face interiors---corner-only sampling under-reports thin objects); project each; cast a ray from the camera; the point is visible if the first hit is the target. \\
bbox fraction & Area of the target's projected bounding box within the frame. \\
view angle & Angle between the viewing ray to the target and the target's surface normal ($0^\circ$ = square-on). \\
px per unit & Pixels spanned by a unit length at the target (text row height, character cell, inch, centimeter): the legibility proxy. \\
clearance & Signed distance from a wire's surface (beveled mesh) to the roof plane; $\le 0$ is contact. \\
apparent reading & For a tape, the blade point that projects onto the module's front-bottom edge from the camera (\S\ref{sec:bench}). \\
skyline / own-roof occlusion & For \env{horizon\_peak} frames: per-column test of the true skyline (panorama tree line or the house's own silhouette) against the frame; ray-cast occlusion by the house's own roof. \\
scene state & A property the scene was built with, read directly: whether the lag bolt and the flashing exist at the attachment. \\
\midrule
blur at target & Circle of confusion from focus error, aperture and focal length, plus motion-blur displacement of the target between shutter poses (px); cross-checked by a Laplacian on the target crop. \\
text contrast / blown fraction & Inside the label mask: Weber contrast of the ink level against the paper level, $(\text{paper}-\text{ink})/\text{paper}$, and the fraction of saturated pixels. \\
paper luminance & The paper level of the label in linear luminance: a dark label is unreadable whatever its contrast. \\
target pixels / edge contrast & Pixels covered by the target (a wire, a module edge) and the mean luminance step across its boundary to the surrounding pixels. \\
\bottomrule
\end{tabular}
\end{table}

\section{Prompt Examples}
\label{app:prompts}
Each item is one query with the frame as the image. Every per-frame prompt has the same frame, with the technician (a solar installer, an electrician or a vehicle-intake technician), the \emph{requirement}, the \emph{question} and the \emph{keys} filled in per environment (Table~\ref{tab:prompt_parts}):

\begin{quote}\small\ttfamily\raggedright You are reviewing a photo taken by <technician> to document their work; the photo criteria will be used to prove the work was carried out correctly. The photo we need: <requirement> First: is this photo adequate -- does it satisfy the requirement as a reviewer would judge it? Then answer the documentation question. <question> Answer with exactly one JSON object and nothing else -- no prose, no code fence -- with exactly these keys: "adequate": true or false (a JSON boolean): whether the photo is adequate; <keys, each with its grammar in words>. Use these values exactly as written; a value in any other form is scored as wrong. Answer every key even if you judge the photo inadequate: give your best reading of what is visible.\end{quote}
For \env{horizon\_peak}, whose requirement is set-level, the frame instead asks whether this one photo belongs in such a set. The full prompt for \env{tape\_gap}:

\begin{quote}\small\ttfamily\raggedright You are reviewing a photo taken by a solar installer to document their work; the photo criteria will be used to prove the work was carried out correctly. The photo we need: Tape measure hooked at the roof surface and standing up to the module frame; the number where the frame crosses the tape must be legible, and the tape should be seated flush (perpendicular to the roof). First: is this photo adequate -- does it satisfy the requirement as a reviewer would judge it? Then answer the documentation question. What number (inches) is on the tape where the module frame crosses it, what is your estimate of the true perpendicular module-to-roof gap in inches, and is the tape leaning rather than standing flush? Answer with exactly one JSON object and nothing else -- no prose, no code fence -- with exactly these keys: "adequate": true or false (a JSON boolean): whether the photo is adequate; "reading\_in": a number (JSON number, no units, no quotes); "gap\_in": a number (JSON number, no units, no quotes); "tilted": true or false (a JSON boolean). Use these values exactly as written; a value in any other form is scored as wrong. Answer every key even if you judge the photo inadequate: give your best reading of what is visible.\end{quote}

\begin{table}[h]\centering\scriptsize\caption{The per-environment parts of the prompt.}\label{tab:prompt_parts}
\setlength{\tabcolsep}{3pt}\begin{tabularx}{\textwidth}{@{}l >{\hsize=1.4\hsize\raggedright\arraybackslash}X >{\hsize=.95\hsize\raggedright\arraybackslash}X >{\hsize=.65\hsize\raggedright\arraybackslash}X@{}}\toprule
Environment & Requirement & Question & Keys \\ \midrule
\env{rail\_attachment} & The L-foot, its flashing and the lag bolt, close enough to judge the attachment. Reject wide array shots where the foot is a speck. & At this roof attachment, is a lag bolt present on the L-foot, and is flashing installed under it? & \texttt{lag\_present}: bool; \texttt{flashing\_present}: bool \\
\env{tape\_gap} & Tape measure hooked at the roof surface and standing up to the module frame; the number where the frame crosses the tape must be legible, and the tape should be seated flush (perpendicular to the roof). & What number (inches) is on the tape where the module frame crosses it, what is your estimate of the true perpendicular module-to-roof gap in inches, and is the tape leaning rather than standing flush? & \texttt{reading\_in}: number; \texttt{gap\_in}: number; \texttt{tilted}: bool \\
\env{wire\_adequacy} & Photograph the underside of each array close enough to judge the wiring: lead wires and trunk cable visible, so a reviewer can tell whether any wire touches the roof plane or has a hard bend. Reject photos that do not clearly show the wiring under the modules. & Is any wire or cable touching the roof surface (the shingles)? & \texttt{touching}: bool \\
\env{slope\_gauge} & A digital angle gauge placed directly on the module glass or frame, aligned with the slope, showing its degree reading; the display must be legible with the module it sits on visible in the photo. Reject a gauge placed on bare roof instead of the module, and a shot where the display is unreadable or washed out. & What does the angle gauge read, in degrees, and is it sitting on the module (glass or frame) or on bare roof? & \texttt{reading\_deg}: number; \texttt{on\_module}: bool \\
\env{equipment\_label} & Close-up of the equipment data label — on the back of an installed module, on a microinverter under the array, or on the inverter — with the model number, the serial number and its barcode all legible, square-on enough to read. Some of the equipment or mounting should be in view so it is clearly a label on site, not a stock image. Reject a shot where the small text is a blur, the barcode is cut off or washed out by glare. & Read the equipment data label: brand, model number and serial number, and say what the label is on. & \texttt{brand}: string; \texttt{model}: string; \texttt{serial}: string; \texttt{on}: module/microinverter/inverter \\
\env{horizon\_peak} & A set of photos taken from the roof peak for the shading analysis: turning in place, each photo shows the skyline in that direction, with sky above it and the surroundings below it, and a little of this house's own roof along the bottom as proof of where it was taken. Together the photos should go most of the way around: four or more directions with no large gap. Reject photos taken from below the peak where the roof blocks the view, photos with no roof in them at all, and sets that leave a large gap. & Is the horizon visible in this frame and roughly level, and does this house's own roof hide part of the far view? & \texttt{horizon\_visible}: bool; \texttt{level}: bool; \texttt{own\_roof\_blocks}: bool \\
\env{damage\_view} & Square onto the damaged area from about an arm's length, with the whole damaged region and part of the adjacent undamaged panel in frame, so a reviewer can judge which panel, how far the damage extends and how severe it is. Reject an oblique or distant shot where the damage is a smudge, and a tight crop with no panel context. & What kind of damage is this, and how severe is it? Severity by the size of the damaged area: minor = under about 10 x 10 cm (100 cm2), moderate = 100 to 500 cm2 (up to about 22 x 22 cm), severe = larger than that. & \texttt{damage\_types}: list of dent/scratch/rust; \texttt{severity}: minor/moderate/severe \\
\env{car\_view} & Straight-on to the requested face of the vehicle (or perpendicular to its flank) from about two car lengths, full width or length of the car visible, both corners in frame. Reject three-quarter angles, frames cropping a corner, and any other car in the lot standing in for it. & Which side of the vehicle is this (front, rear, driver side, passenger side)? Also give the vehicle's body style. & \texttt{side}: front/rear/driver/passenger; \texttt{body\_style}: one of 10 body styles \\
\env{car\_plate} & Straight-on close-up of the mounted plate on the vehicle, filling most of the frame, characters legible end to end. Reject a wide shot of the whole car where the plate is a smudge, and the plate of any other car in the lot. & Read the license plate: its characters and the state named on it. & \texttt{plate}: string; \texttt{state}: string \\
\env{ro\_board} & The repair-order board (RO number, VIN, date) readable in the same frame as the vehicle: close and square enough that the writing reads, with the car clearly visible behind or beside it. Reject a wide shot where the board text is a smudge, and a board close-up with no vehicle in it. & Read the repair-order board: the RO number and the VIN. Also give the body style of the vehicle in the frame. & \texttt{ro}: string; \texttt{vin}: string; \texttt{body\_style}: one of 10 body styles \\
\env{breaker\_label} & Tight on the label card or handwritten directory beside the breaker rows so every circuit name and its amperage read cleanly. Reject a full-panel shot where the label text is illegible, and a bare breaker column with no labels in frame. & Read the whole circuit directory card: the main breaker amperage and every circuit's name and amperage. & \texttt{main\_amps}: number; \texttt{entries}: list of \{name, amps\} \\
\bottomrule\end{tabularx}\end{table}

\FloatBarrier

\section{Evaluation details}
\label{app:static}
\paragraph{Generation.} Frames that fail the render sanity check (camera inside a mesh, black or blown frame, a car mesh intersecting scenery) are excluded before sampling. Because the scorer accepts only 5--39\% of random field poses (Table~\ref{tab:envs}), the miss rate of a model cannot be estimated per environment from 100 random frames alone. Thus, for the seven environments where random frames are rarely acceptable we add a \emph{positive stratum} of 50 scorer-acceptable frames obtained by rejection sampling from the same pose distribution. In \env{ro\_board}, 21 random and 8 positive-stratum frames were withdrawn after generation because of a defect in the board texture. The false-accept rate is measured on the random frames only.

\paragraph{Test conditions.} Each frame is one query: the reviewer framing and the checklist requirement, then one JSON object whose first key is the acceptability verdict and whose remaining keys are the documentation answer in a stated grammar, answered whatever the verdict (the effect of this order is tested below). Acceptability is scored against the scorer's verdict; documentation against the ground truth per field (exact match after normalization, which for text upper-cases and keeps only letters and digits; for the breaker directory, the fraction of the card's rows matched by a reply row with the same name and amperage, in any order, each reply row used once, extra rows not penalized; tolerances of $\nicefrac{1}{8}$\,in, $\nicefrac{1}{4}$\,in and $0.2^\circ$ for the tape reading, the gap and the gauge reading).

\paragraph{Does the order of the answer matter?}\label{app:order} We re-ask \orderN{} items from \env{tape\_gap}, \env{slope\_gauge} and \env{rail\_attachment} under three more prompts that differ from the benchmark's only as stated: the verdict as the \emph{last} key, with the two instructions swapped to match; the documentation question \emph{alone}, with the same framing and requirement; and the verdict alone. The benchmark's own prompt is re-run to give the run-to-run noise. We ran \orderFrontierModels{} of the \nFrontier{} models on the cost frontiers of Fig.~\ref{fig:static_cost} and \the\numexpr\orderModels-\orderFrontierModels\relax{} others (Table~\ref{tab:order_ablation}). Gemini 3.1 Flash Lite reports fewer defects when the verdict is moved last, and about half as many, in all three environments, when no verdict is asked; Gemma 4 31B reports fewer missing lag bolts and flashings in \env{rail\_attachment} when no verdict is asked. No other model changes beyond the re-run noise, and no model's documentation score on these items is lower without the verdict.

\begin{table}[t]\centering\scriptsize\setlength{\tabcolsep}{3.5pt}\caption{Verdict-order ablation: 150 items from \env{tape\_gap}, \env{slope\_gauge} and \env{rail\_attachment} (25 per environment from the random frames and 25 from the positive stratum), re-asked under four prompts: the benchmark's (\emph{first}: the verdict is the first key), verdict \emph{last}, documentation \emph{alone} (no verdict asked) and the verdict alone. A \emph{defect claim} is the documentation boolean in the direction the requirement rejects (tape tilted; gauge not on the module; lag bolt or flashing absent). Paired columns count items whose claim is present under \emph{first} only / under the other prompt only, with the exact McNemar $p$; \emph{rerun} compares \emph{first} with the same prompt's record in the main run (the noise floor). \emph{Rejects}: frames rejected under \emph{first} and when the verdict is asked alone. $^\ddagger$ on a cost frontier of Fig.~\ref{fig:static_cost}. One sample per prompt; provider-default thinking.}\label{tab:order_ablation}
\begin{tabular}{@{}l rrr rrr rr@{}}\toprule
 & \multicolumn{3}{c}{defect claims (of 150)} & \multicolumn{3}{c}{paired change in claims vs.\ \emph{first}} & \multicolumn{2}{c}{rejects} \\ \cmidrule(lr){2-4}\cmidrule(lr){5-7}\cmidrule(l){8-9}
Model & first & last & alone & last & alone ($p$) & rerun & first & alone ($p$) \\ \midrule
Gemini 3.8 Flash$^\ddagger$ & 82 & 85 & 83 & 4/7 & 7/8 (1.00) & 8/5 & 96 & 92 (.60) \\
Gemini 3.5 Flash & 94 & 97 & 88 & 7/10 & 12/6 (.24) & 7/6 & 104 & 108 (.50) \\
Gemini 3.1 Flash Lite & 125 & 103 & 66 & 27/5 & 60/1 ($<$.001) & 10/6 & 125 & 133 (.13) \\
Gemma 4 31B$^\ddagger$ & 92 & 96 & 78 & 5/9 & 18/4 (.004) & 8/7 & 101 & 107 (.36) \\
GPT-6 Astra$^\ddagger$ & 58 & 53 & 56 & 8/3 & 9/7 (.80) & 9/6 & 76 & 73 (.65) \\
GPT-5.6 Sol & 63 & 70 & 61 & 7/14 & 12/10 (.83) & 14/11 & 81 & 97 (.002) \\
Claude Opus 5 & 100 & 98 & 95 & 8/6 & 15/10 (.42) & 6/14 & 113 & 141 ($<$.001) \\
Claude Sonnet 5 & 119 & 118 & 110 & 8/7 & 18/9 (.12) & 4/4 & 122 & 120 (.84) \\
GLM-5.3 Flash$^\ddagger$ & 82 & 74 & 76 & 19/11 & 17/11 (.34) & 12/12 & 91 & 105 (.03) \\
\bottomrule\end{tabular}\end{table}

\paragraph{Detailed results.} Table~\ref{tab:static_v2_totals} gives, per model on the same 1,378 frames, the false-accept and miss rates, acceptability precision, recall and F1 with a 95\% interval, the documentation score with its interval, the fraction of answers in grammar, the cost of the set and the mean thinking tokens per item. Table~\ref{tab:static_v2_paired} gives the paired difference in F1 and in documentation score for every pair of models, and whether its interval excludes zero (every model saw the same items, so the bootstrap resamples the items once and scores both models on that resample). Table~\ref{tab:static_v2_prices} lists the token prices with their sources and dates. Table~\ref{tab:static_v2_f1} breaks acceptability F1 down by environment and Table~\ref{tab:static_v2_fa} the false-accept rate; Table~\ref{tab:field_scores_full} gives the documentation score per field.
False-accept and miss rates are against the scorer's verdict; the documentation score is the mean, over the frames the scorer accepts, of each frame's mean field score (fields as above; the breaker main amperage and rows count half each). Costs use the prices of Table~\ref{tab:static_v2_prices} (the list API price of the model's developer, read from the developer's own pricing page on the date shown, whichever host served our calls), times the tokens the API reported: every prompt token at the full input rate, thinking tokens at the output rate. Google lists a per-token price for Gemma 4 26B only, so Gemma 4 26B-A4B and Gemma 4 31B are both priced at that rate.
\begin{table}[t]\centering\scriptsize\setlength{\tabcolsep}{3pt}\caption{Detailed statistics. False-accept rate (random frames the scorer rejects that the model accepts) and miss rate (scorer-acceptable frames, random and positive strata, that the model rejects); acceptability precision, recall and F1 against the scorer (precision on the 934 random frames, recall including the positive strata), with a 95\% interval on F1; the documentation score on scorer-acceptable frames (mean per-field score) with its 95\% interval; the fraction of answers in grammar; the cost of the set at the price in Table~\ref{tab:static_v2_prices} (thinking at the output rate); mean thinking tokens per item. Below the rule: Gemma 4 E4B, Qwen3.5-4B and LFM2.5-VL-3B, which we served ourselves; their developers list no per-token price, so they have no cost. They sit below the same rule in the per-environment tables and are in every comparison except the cost frontier, which needs a price. Intervals: stratified bootstrap over items, 1,000 resamples, the random frames and each positive stratum resampled separately. Replies cut by the output cap: Qwen 3.8 27B 6, Qwen 3.5 397B-A17B 10, Kimi K3 6, DeepSeek V4.1 Flash 71, GLM-5.3 Flash 14.}\label{tab:static_v2_totals}
\begin{tabular}{@{}lrrrrrrrrr@{}}\toprule
Model & FA (\%) & miss (\%) & P & R & F1 [95\% CI] & doc.\ [95\% CI] & gram.\ (\%) & cost (\$) & think \\ \midrule
Gemini 3.8 Flash & 11 & 38 & 0.48 & 0.62 & 0.54 {\tiny[0.48, 0.59]} & 0.84 {\tiny[0.82, 0.86]} & 100 & 7.16 & 1083 \\
Gemini 3.5 Flash & 13 & 34 & 0.49 & 0.66 & 0.56 {\tiny[0.51, 0.62]} & 0.80 {\tiny[0.78, 0.83]} & 97 & 7.36 & 1114 \\
Gemini 3.1 Flash Lite & 16 & 42 & 0.39 & 0.58 & 0.47 {\tiny[0.41, 0.52]} & 0.73 {\tiny[0.70, 0.75]} & 98 & 0.73 & 0 \\
Gemma 4 31B & 15 & 37 & 0.44 & 0.63 & 0.52 {\tiny[0.46, 0.57]} & 0.79 {\tiny[0.77, 0.81]} & 100 & 0.70 & 677 \\
Gemma 4 26B-A4B & 17 & 36 & 0.39 & 0.64 & 0.48 {\tiny[0.42, 0.53]} & 0.77 {\tiny[0.75, 0.79]} & 99 & 1.24 & 1325 \\
GPT-6 Astra & 12 & 32 & 0.55 & 0.68 & 0.61 {\tiny[0.56, 0.66]} & 0.83 {\tiny[0.81, 0.85]} & 100 & 30.05 & 165 \\
GPT-6 Sol & 13 & 39 & 0.50 & 0.61 & 0.55 {\tiny[0.50, 0.60]} & 0.78 {\tiny[0.75, 0.80]} & 100 & 8.13 & 320 \\
GPT-6 Luna & 9 & 59 & 0.46 & 0.41 & 0.43 {\tiny[0.38, 0.48]} & 0.62 {\tiny[0.60, 0.65]} & 100 & 0.69 & 751 \\
GPT-5.6 Sol & 16 & 31 & 0.49 & 0.69 & 0.57 {\tiny[0.52, 0.62]} & 0.77 {\tiny[0.75, 0.80]} & 100 & 17.65 & 368 \\
GPT-5.6 Terra & 17 & 30 & 0.45 & 0.70 & 0.55 {\tiny[0.49, 0.60]} & 0.76 {\tiny[0.74, 0.78]} & 100 & 8.49 & 282 \\
GPT-5.6 Luna & 16 & 38 & 0.46 & 0.62 & 0.53 {\tiny[0.47, 0.58]} & 0.71 {\tiny[0.69, 0.74]} & 100 & 1.18 & 482 \\
Claude Fable 5.1 & 12 & 37 & 0.51 & 0.63 & 0.56 {\tiny[0.51, 0.61]} & 0.83 {\tiny[0.81, 0.85]} & 100 & 41.35 & 267 \\
Claude Opus 5.5 & 14 & 33 & 0.51 & 0.67 & 0.58 {\tiny[0.52, 0.62]} & 0.83 {\tiny[0.80, 0.85]} & 100 & 16.36 & 263 \\
Claude Opus 5 & 11 & 43 & 0.54 & 0.57 & 0.56 {\tiny[0.50, 0.60]} & 0.75 {\tiny[0.73, 0.78]} & 100 & 23.32 & 345 \\
Claude Sonnet 5 & 12 & 49 & 0.45 & 0.51 & 0.48 {\tiny[0.42, 0.53]} & 0.73 {\tiny[0.70, 0.75]} & 100 & 4.80 & 19 \\
Qwen 3.8 Max & 23 & 22 & 0.43 & 0.78 & 0.55 {\tiny[0.50, 0.60]} & 0.82 {\tiny[0.80, 0.84]} & 100 & 38.19 & 4231 \\
Qwen 3.8 27B & 15 & 32 & 0.45 & 0.68 & 0.54 {\tiny[0.48, 0.60]} & 0.74 {\tiny[0.72, 0.76]} & 99 & 12.36 & 2763 \\
Qwen 3.5 397B-A17B & 11 & 39 & 0.49 & 0.61 & 0.55 {\tiny[0.49, 0.60]} & 0.76 {\tiny[0.74, 0.79]} & 99 & 15.03 & 2799 \\
Grok 4.6 & 19 & 28 & 0.43 & 0.72 & 0.54 {\tiny[0.49, 0.59]} & 0.78 {\tiny[0.75, 0.80]} & 100 & 24.88 & 2634 \\
Kimi K3 & 20 & 30 & 0.42 & 0.70 & 0.53 {\tiny[0.48, 0.58]} & 0.77 {\tiny[0.75, 0.80]} & 99 & 39.31 & 1582 \\
Muse Spark 1.3 & 11 & 38 & 0.49 & 0.62 & 0.55 {\tiny[0.49, 0.60]} & 0.79 {\tiny[0.76, 0.81]} & 100 & 13.22 & 1825 \\
DeepSeek V4.1 Flash & 18 & 32 & 0.39 & 0.68 & 0.50 {\tiny[0.44, 0.55]} & 0.68 {\tiny[0.65, 0.71]} & 95 & 8.67 & 4903 \\
GLM-5.3 Flash & 14 & 32 & 0.49 & 0.68 & 0.57 {\tiny[0.51, 0.62]} & 0.75 {\tiny[0.72, 0.77]} & 99 & 1.30 & 1460 \\
\midrule
Gemma 4 E4B & 33 & 30 & 0.30 & 0.70 & 0.42 {\tiny[0.37, 0.47]} & 0.69 {\tiny[0.66, 0.71]} & 98 & -- & 740 \\
Qwen3.5-4B & 18 & 50 & 0.31 & 0.50 & 0.39 {\tiny[0.33, 0.44]} & 0.66 {\tiny[0.64, 0.69]} & 95 & -- & 0 \\
LFM2.5-VL-3B & 40 & 21 & 0.30 & 0.79 & 0.43 {\tiny[0.39, 0.48]} & 0.68 {\tiny[0.66, 0.71]} & 99 & -- & 0 \\
\bottomrule\end{tabular}\end{table}

\begin{table}[p]\centering\scriptsize\setlength{\tabcolsep}{2.5pt}\caption{Paired differences between models on the same 1,378 frames, acceptability F1 (top) and documentation score (bottom). Models are sorted by the metric (its value beside the name); a cell is the column model's value minus the row model's, both as printed, in hundredths. \textbf{Bold}: the paired 95\% interval of the unrounded difference excludes zero; \textcolor{gray}{gray}: it does not (160 of 325 pairs for F1, 76 for documentation). $^\S$ no developer list price (served by us). Intervals: a stratified bootstrap that resamples the items once and scores both models on the same resample (2000 resamples).}\label{tab:static_v2_paired}
\begin{tabular}{@{}lrrrrrrrrrrrrrrrrrrrrrrrrrr@{}}\toprule
F1 & & \rotatebox{90}{GPT-6 Astra} & \rotatebox{90}{Claude Opus 5.5} & \rotatebox{90}{GPT-5.6 Sol} & \rotatebox{90}{GLM-5.3 Flash} & \rotatebox{90}{Gemini 3.5 Flash} & \rotatebox{90}{Claude Fable 5.1} & \rotatebox{90}{Claude Opus 5} & \rotatebox{90}{Qwen 3.8 Max} & \rotatebox{90}{GPT-6 Sol} & \rotatebox{90}{Muse Spark 1.3} & \rotatebox{90}{Qwen 3.5 397B-A17B} & \rotatebox{90}{GPT-5.6 Terra} & \rotatebox{90}{Qwen 3.8 27B} & \rotatebox{90}{Grok 4.6} & \rotatebox{90}{Gemini 3.8 Flash} & \rotatebox{90}{Kimi K3} & \rotatebox{90}{GPT-5.6 Luna} & \rotatebox{90}{Gemma 4 31B} & \rotatebox{90}{DeepSeek V4.1 Flash} & \rotatebox{90}{Gemma 4 26B-A4B} & \rotatebox{90}{Claude Sonnet 5} & \rotatebox{90}{Gemini 3.1 Flash Lite} & \rotatebox{90}{LFM2.5-VL-3B$^\S$} & \rotatebox{90}{GPT-6 Luna} & \rotatebox{90}{Gemma 4 E4B$^\S$} \\ \midrule
GPT-6 Astra & 0.61 &  &  &  &  &  &  &  &  &  &  &  &  &  &  &  &  &  &  &  &  &  &  &  &  &  \\
Claude Opus 5.5 & 0.58 & \textcolor{gray}{3} &  &  &  &  &  &  &  &  &  &  &  &  &  &  &  &  &  &  &  &  &  &  &  &  \\
GPT-5.6 Sol & 0.57 & \textbf{4} & \textcolor{gray}{1} &  &  &  &  &  &  &  &  &  &  &  &  &  &  &  &  &  &  &  &  &  &  &  \\
GLM-5.3 Flash & 0.57 & \textcolor{gray}{4} & \textcolor{gray}{1} & \textcolor{gray}{0} &  &  &  &  &  &  &  &  &  &  &  &  &  &  &  &  &  &  &  &  &  &  \\
Gemini 3.5 Flash & 0.56 & \textcolor{gray}{5} & \textcolor{gray}{2} & \textcolor{gray}{1} & \textcolor{gray}{1} &  &  &  &  &  &  &  &  &  &  &  &  &  &  &  &  &  &  &  &  &  \\
Claude Fable 5.1 & 0.56 & \textbf{5} & \textcolor{gray}{2} & \textcolor{gray}{1} & \textcolor{gray}{1} & \textcolor{gray}{0} &  &  &  &  &  &  &  &  &  &  &  &  &  &  &  &  &  &  &  &  \\
Claude Opus 5 & 0.56 & \textbf{5} & \textcolor{gray}{2} & \textcolor{gray}{1} & \textcolor{gray}{1} & \textcolor{gray}{0} & \textcolor{gray}{0} &  &  &  &  &  &  &  &  &  &  &  &  &  &  &  &  &  &  &  \\
Qwen 3.8 Max & 0.55 & \textbf{6} & \textcolor{gray}{3} & \textcolor{gray}{2} & \textcolor{gray}{2} & \textcolor{gray}{1} & \textcolor{gray}{1} & \textcolor{gray}{1} &  &  &  &  &  &  &  &  &  &  &  &  &  &  &  &  &  &  \\
GPT-6 Sol & 0.55 & \textbf{6} & \textcolor{gray}{3} & \textcolor{gray}{2} & \textcolor{gray}{2} & \textcolor{gray}{1} & \textcolor{gray}{1} & \textcolor{gray}{1} & \textcolor{gray}{0} &  &  &  &  &  &  &  &  &  &  &  &  &  &  &  &  &  \\
Muse Spark 1.3 & 0.55 & \textbf{6} & \textcolor{gray}{3} & \textcolor{gray}{2} & \textcolor{gray}{2} & \textcolor{gray}{1} & \textcolor{gray}{1} & \textcolor{gray}{1} & \textcolor{gray}{0} & \textcolor{gray}{0} &  &  &  &  &  &  &  &  &  &  &  &  &  &  &  &  \\
Qwen 3.5 397B-A17B & 0.55 & \textbf{6} & \textcolor{gray}{3} & \textcolor{gray}{2} & \textcolor{gray}{2} & \textcolor{gray}{1} & \textcolor{gray}{1} & \textcolor{gray}{1} & \textcolor{gray}{0} & \textcolor{gray}{0} & \textcolor{gray}{0} &  &  &  &  &  &  &  &  &  &  &  &  &  &  &  \\
GPT-5.6 Terra & 0.55 & \textbf{6} & \textcolor{gray}{3} & \textcolor{gray}{2} & \textcolor{gray}{2} & \textcolor{gray}{1} & \textcolor{gray}{1} & \textcolor{gray}{1} & \textcolor{gray}{0} & \textcolor{gray}{0} & \textcolor{gray}{0} & \textcolor{gray}{0} &  &  &  &  &  &  &  &  &  &  &  &  &  &  \\
Qwen 3.8 27B & 0.54 & \textbf{7} & \textcolor{gray}{4} & \textcolor{gray}{3} & \textcolor{gray}{3} & \textcolor{gray}{2} & \textcolor{gray}{2} & \textcolor{gray}{2} & \textcolor{gray}{1} & \textcolor{gray}{1} & \textcolor{gray}{1} & \textcolor{gray}{1} & \textcolor{gray}{1} &  &  &  &  &  &  &  &  &  &  &  &  &  \\
Grok 4.6 & 0.54 & \textbf{7} & \textcolor{gray}{4} & \textcolor{gray}{3} & \textcolor{gray}{3} & \textcolor{gray}{2} & \textcolor{gray}{2} & \textcolor{gray}{2} & \textcolor{gray}{1} & \textcolor{gray}{1} & \textcolor{gray}{1} & \textcolor{gray}{1} & \textcolor{gray}{1} & \textcolor{gray}{0} &  &  &  &  &  &  &  &  &  &  &  &  \\
Gemini 3.8 Flash & 0.54 & \textbf{7} & \textcolor{gray}{4} & \textcolor{gray}{3} & \textcolor{gray}{3} & \textcolor{gray}{2} & \textcolor{gray}{2} & \textcolor{gray}{2} & \textcolor{gray}{1} & \textcolor{gray}{1} & \textcolor{gray}{1} & \textcolor{gray}{1} & \textcolor{gray}{1} & \textcolor{gray}{0} & \textcolor{gray}{0} &  &  &  &  &  &  &  &  &  &  &  \\
Kimi K3 & 0.53 & \textbf{8} & \textbf{5} & \textbf{4} & \textcolor{gray}{4} & \textcolor{gray}{3} & \textcolor{gray}{3} & \textcolor{gray}{3} & \textcolor{gray}{2} & \textcolor{gray}{2} & \textcolor{gray}{2} & \textcolor{gray}{2} & \textcolor{gray}{2} & \textcolor{gray}{1} & \textcolor{gray}{1} & \textcolor{gray}{1} &  &  &  &  &  &  &  &  &  &  \\
GPT-5.6 Luna & 0.53 & \textbf{8} & \textbf{5} & \textbf{4} & \textcolor{gray}{4} & \textcolor{gray}{3} & \textcolor{gray}{3} & \textcolor{gray}{3} & \textcolor{gray}{2} & \textcolor{gray}{2} & \textcolor{gray}{2} & \textcolor{gray}{2} & \textcolor{gray}{2} & \textcolor{gray}{1} & \textcolor{gray}{1} & \textcolor{gray}{1} & \textcolor{gray}{0} &  &  &  &  &  &  &  &  &  \\
Gemma 4 31B & 0.52 & \textbf{9} & \textbf{6} & \textbf{5} & \textbf{5} & \textbf{4} & \textcolor{gray}{4} & \textcolor{gray}{4} & \textcolor{gray}{3} & \textcolor{gray}{3} & \textcolor{gray}{3} & \textcolor{gray}{3} & \textcolor{gray}{3} & \textcolor{gray}{2} & \textcolor{gray}{2} & \textcolor{gray}{2} & \textcolor{gray}{1} & \textcolor{gray}{1} &  &  &  &  &  &  &  &  \\
DeepSeek V4.1 Flash & 0.50 & \textbf{11} & \textbf{8} & \textbf{7} & \textbf{7} & \textbf{6} & \textbf{6} & \textbf{6} & \textbf{5} & \textbf{5} & \textcolor{gray}{5} & \textbf{5} & \textcolor{gray}{5} & \textbf{4} & \textcolor{gray}{4} & \textcolor{gray}{4} & \textcolor{gray}{3} & \textcolor{gray}{3} & \textcolor{gray}{2} &  &  &  &  &  &  &  \\
Gemma 4 26B-A4B & 0.48 & \textbf{13} & \textbf{10} & \textbf{9} & \textbf{9} & \textbf{8} & \textbf{8} & \textbf{8} & \textbf{7} & \textbf{7} & \textbf{7} & \textbf{7} & \textbf{7} & \textbf{6} & \textbf{6} & \textbf{6} & \textbf{5} & \textcolor{gray}{5} & \textcolor{gray}{4} & \textcolor{gray}{2} &  &  &  &  &  &  \\
Claude Sonnet 5 & 0.48 & \textbf{13} & \textbf{10} & \textbf{9} & \textbf{9} & \textbf{8} & \textbf{8} & \textbf{8} & \textbf{7} & \textbf{7} & \textbf{7} & \textbf{7} & \textbf{7} & \textbf{6} & \textbf{6} & \textbf{6} & \textbf{5} & \textcolor{gray}{5} & \textcolor{gray}{4} & \textcolor{gray}{2} & \textcolor{gray}{0} &  &  &  &  &  \\
Gemini 3.1 Flash Lite & 0.47 & \textbf{14} & \textbf{11} & \textbf{10} & \textbf{10} & \textbf{9} & \textbf{9} & \textbf{9} & \textbf{8} & \textbf{8} & \textbf{8} & \textbf{8} & \textbf{8} & \textbf{7} & \textbf{7} & \textbf{7} & \textbf{6} & \textbf{6} & \textbf{5} & \textcolor{gray}{3} & \textcolor{gray}{1} & \textcolor{gray}{1} &  &  &  &  \\
LFM2.5-VL-3B$^\S$ & 0.43 & \textbf{18} & \textbf{15} & \textbf{14} & \textbf{14} & \textbf{13} & \textbf{13} & \textbf{13} & \textbf{12} & \textbf{12} & \textbf{12} & \textbf{12} & \textbf{12} & \textbf{11} & \textbf{11} & \textbf{11} & \textbf{10} & \textbf{10} & \textbf{9} & \textbf{7} & \textcolor{gray}{5} & \textcolor{gray}{5} & \textcolor{gray}{4} &  &  &  \\
GPT-6 Luna & 0.43 & \textbf{18} & \textbf{15} & \textbf{14} & \textbf{14} & \textbf{13} & \textbf{13} & \textbf{13} & \textbf{12} & \textbf{12} & \textbf{12} & \textbf{12} & \textbf{12} & \textbf{11} & \textbf{11} & \textbf{11} & \textbf{10} & \textbf{10} & \textbf{9} & \textbf{7} & \textcolor{gray}{5} & \textcolor{gray}{5} & \textcolor{gray}{4} & \textcolor{gray}{0} &  &  \\
Gemma 4 E4B$^\S$ & 0.42 & \textbf{19} & \textbf{16} & \textbf{15} & \textbf{15} & \textbf{14} & \textbf{14} & \textbf{14} & \textbf{13} & \textbf{13} & \textbf{13} & \textbf{13} & \textbf{13} & \textbf{12} & \textbf{12} & \textbf{12} & \textbf{11} & \textbf{11} & \textbf{10} & \textbf{8} & \textbf{6} & \textbf{6} & \textcolor{gray}{5} & \textcolor{gray}{1} & \textcolor{gray}{1} &  \\
Qwen3.5-4B$^\S$ & 0.39 & \textbf{22} & \textbf{19} & \textbf{18} & \textbf{18} & \textbf{17} & \textbf{17} & \textbf{17} & \textbf{16} & \textbf{16} & \textbf{16} & \textbf{16} & \textbf{16} & \textbf{15} & \textbf{15} & \textbf{15} & \textbf{14} & \textbf{14} & \textbf{13} & \textbf{11} & \textbf{9} & \textbf{9} & \textbf{8} & \textbf{4} & \textcolor{gray}{4} & \textcolor{gray}{3} \\
\bottomrule\end{tabular}

\vspace{1.5em}
\begin{tabular}{@{}lrrrrrrrrrrrrrrrrrrrrrrrrrr@{}}\toprule
Documentation & & \rotatebox{90}{Gemini 3.8 Flash} & \rotatebox{90}{Claude Opus 5.5} & \rotatebox{90}{GPT-6 Astra} & \rotatebox{90}{Claude Fable 5.1} & \rotatebox{90}{Qwen 3.8 Max} & \rotatebox{90}{Gemini 3.5 Flash} & \rotatebox{90}{Gemma 4 31B} & \rotatebox{90}{Muse Spark 1.3} & \rotatebox{90}{GPT-6 Sol} & \rotatebox{90}{Grok 4.6} & \rotatebox{90}{Kimi K3} & \rotatebox{90}{GPT-5.6 Sol} & \rotatebox{90}{Gemma 4 26B-A4B} & \rotatebox{90}{Qwen 3.5 397B-A17B} & \rotatebox{90}{GPT-5.6 Terra} & \rotatebox{90}{Claude Opus 5} & \rotatebox{90}{GLM-5.3 Flash} & \rotatebox{90}{Qwen 3.8 27B} & \rotatebox{90}{Gemini 3.1 Flash Lite} & \rotatebox{90}{Claude Sonnet 5} & \rotatebox{90}{GPT-5.6 Luna} & \rotatebox{90}{Gemma 4 E4B$^\S$} & \rotatebox{90}{LFM2.5-VL-3B$^\S$} & \rotatebox{90}{DeepSeek V4.1 Flash} & \rotatebox{90}{Qwen3.5-4B$^\S$} \\ \midrule
Gemini 3.8 Flash & 0.84 &  &  &  &  &  &  &  &  &  &  &  &  &  &  &  &  &  &  &  &  &  &  &  &  &  \\
Claude Opus 5.5 & 0.83 & \textcolor{gray}{1} &  &  &  &  &  &  &  &  &  &  &  &  &  &  &  &  &  &  &  &  &  &  &  &  \\
GPT-6 Astra & 0.83 & \textcolor{gray}{1} & \textcolor{gray}{0} &  &  &  &  &  &  &  &  &  &  &  &  &  &  &  &  &  &  &  &  &  &  &  \\
Claude Fable 5.1 & 0.83 & \textcolor{gray}{1} & \textcolor{gray}{0} & \textcolor{gray}{0} &  &  &  &  &  &  &  &  &  &  &  &  &  &  &  &  &  &  &  &  &  &  \\
Qwen 3.8 Max & 0.82 & \textcolor{gray}{2} & \textcolor{gray}{1} & \textcolor{gray}{1} & \textcolor{gray}{1} &  &  &  &  &  &  &  &  &  &  &  &  &  &  &  &  &  &  &  &  &  \\
Gemini 3.5 Flash & 0.80 & \textbf{4} & \textbf{3} & \textcolor{gray}{3} & \textbf{3} & \textcolor{gray}{2} &  &  &  &  &  &  &  &  &  &  &  &  &  &  &  &  &  &  &  &  \\
Gemma 4 31B & 0.79 & \textbf{5} & \textbf{4} & \textbf{4} & \textbf{4} & \textbf{3} & \textcolor{gray}{1} &  &  &  &  &  &  &  &  &  &  &  &  &  &  &  &  &  &  &  \\
Muse Spark 1.3 & 0.79 & \textbf{5} & \textbf{4} & \textbf{4} & \textbf{4} & \textbf{3} & \textcolor{gray}{1} & \textcolor{gray}{0} &  &  &  &  &  &  &  &  &  &  &  &  &  &  &  &  &  &  \\
GPT-6 Sol & 0.78 & \textbf{6} & \textbf{5} & \textbf{5} & \textbf{5} & \textbf{4} & \textbf{2} & \textcolor{gray}{1} & \textcolor{gray}{1} &  &  &  &  &  &  &  &  &  &  &  &  &  &  &  &  &  \\
Grok 4.6 & 0.78 & \textbf{6} & \textbf{5} & \textbf{5} & \textbf{5} & \textbf{4} & \textbf{2} & \textcolor{gray}{1} & \textcolor{gray}{1} & \textcolor{gray}{0} &  &  &  &  &  &  &  &  &  &  &  &  &  &  &  &  \\
Kimi K3 & 0.77 & \textbf{7} & \textbf{6} & \textbf{6} & \textbf{6} & \textbf{5} & \textbf{3} & \textcolor{gray}{2} & \textcolor{gray}{2} & \textcolor{gray}{1} & \textcolor{gray}{1} &  &  &  &  &  &  &  &  &  &  &  &  &  &  &  \\
GPT-5.6 Sol & 0.77 & \textbf{7} & \textbf{6} & \textbf{6} & \textbf{6} & \textbf{5} & \textbf{3} & \textcolor{gray}{2} & \textcolor{gray}{2} & \textcolor{gray}{1} & \textcolor{gray}{1} & \textcolor{gray}{0} &  &  &  &  &  &  &  &  &  &  &  &  &  &  \\
Gemma 4 26B-A4B & 0.77 & \textbf{7} & \textbf{6} & \textbf{6} & \textbf{6} & \textbf{5} & \textbf{3} & \textbf{2} & \textcolor{gray}{2} & \textcolor{gray}{1} & \textcolor{gray}{1} & \textcolor{gray}{0} & \textcolor{gray}{0} &  &  &  &  &  &  &  &  &  &  &  &  &  \\
Qwen 3.5 397B-A17B & 0.76 & \textbf{8} & \textbf{7} & \textbf{7} & \textbf{7} & \textbf{6} & \textbf{4} & \textbf{3} & \textbf{3} & \textcolor{gray}{2} & \textcolor{gray}{2} & \textcolor{gray}{1} & \textcolor{gray}{1} & \textcolor{gray}{1} &  &  &  &  &  &  &  &  &  &  &  &  \\
GPT-5.6 Terra & 0.76 & \textbf{8} & \textbf{7} & \textbf{7} & \textbf{7} & \textbf{6} & \textbf{4} & \textbf{3} & \textbf{3} & \textcolor{gray}{2} & \textcolor{gray}{2} & \textcolor{gray}{1} & \textcolor{gray}{1} & \textcolor{gray}{1} & \textcolor{gray}{0} &  &  &  &  &  &  &  &  &  &  &  \\
Claude Opus 5 & 0.75 & \textbf{9} & \textbf{8} & \textbf{8} & \textbf{8} & \textbf{7} & \textbf{5} & \textbf{4} & \textbf{4} & \textcolor{gray}{3} & \textcolor{gray}{3} & \textcolor{gray}{2} & \textcolor{gray}{2} & \textcolor{gray}{2} & \textcolor{gray}{1} & \textcolor{gray}{1} &  &  &  &  &  &  &  &  &  &  \\
GLM-5.3 Flash & 0.75 & \textbf{9} & \textbf{8} & \textbf{8} & \textbf{8} & \textbf{7} & \textbf{5} & \textbf{4} & \textbf{4} & \textbf{3} & \textbf{3} & \textbf{2} & \textcolor{gray}{2} & \textcolor{gray}{2} & \textcolor{gray}{1} & \textcolor{gray}{1} & \textcolor{gray}{0} &  &  &  &  &  &  &  &  &  \\
Qwen 3.8 27B & 0.74 & \textbf{10} & \textbf{9} & \textbf{9} & \textbf{9} & \textbf{8} & \textbf{6} & \textbf{5} & \textbf{5} & \textbf{4} & \textbf{4} & \textbf{3} & \textbf{3} & \textbf{3} & \textcolor{gray}{2} & \textcolor{gray}{2} & \textcolor{gray}{1} & \textcolor{gray}{1} &  &  &  &  &  &  &  &  \\
Gemini 3.1 Flash Lite & 0.73 & \textbf{11} & \textbf{10} & \textbf{10} & \textbf{10} & \textbf{9} & \textbf{7} & \textbf{6} & \textbf{6} & \textbf{5} & \textbf{5} & \textbf{4} & \textbf{4} & \textbf{4} & \textbf{3} & \textbf{3} & \textbf{2} & \textcolor{gray}{2} & \textcolor{gray}{1} &  &  &  &  &  &  &  \\
Claude Sonnet 5 & 0.73 & \textbf{11} & \textbf{10} & \textbf{10} & \textbf{10} & \textbf{9} & \textbf{7} & \textbf{6} & \textbf{6} & \textbf{5} & \textbf{5} & \textbf{4} & \textbf{4} & \textbf{4} & \textbf{3} & \textbf{3} & \textbf{2} & \textcolor{gray}{2} & \textcolor{gray}{1} & \textcolor{gray}{0} &  &  &  &  &  &  \\
GPT-5.6 Luna & 0.71 & \textbf{13} & \textbf{12} & \textbf{12} & \textbf{12} & \textbf{11} & \textbf{9} & \textbf{8} & \textbf{8} & \textbf{7} & \textbf{7} & \textbf{6} & \textbf{6} & \textbf{6} & \textbf{5} & \textbf{5} & \textbf{4} & \textbf{4} & \textbf{3} & \textcolor{gray}{2} & \textcolor{gray}{2} &  &  &  &  &  \\
Gemma 4 E4B$^\S$ & 0.69 & \textbf{15} & \textbf{14} & \textbf{14} & \textbf{14} & \textbf{13} & \textbf{11} & \textbf{10} & \textbf{10} & \textbf{9} & \textbf{9} & \textbf{8} & \textbf{8} & \textbf{8} & \textbf{7} & \textbf{7} & \textbf{6} & \textbf{6} & \textbf{5} & \textbf{4} & \textbf{4} & \textcolor{gray}{2} &  &  &  &  \\
LFM2.5-VL-3B$^\S$ & 0.68 & \textbf{16} & \textbf{15} & \textbf{15} & \textbf{15} & \textbf{14} & \textbf{12} & \textbf{11} & \textbf{11} & \textbf{10} & \textbf{10} & \textbf{9} & \textbf{9} & \textbf{9} & \textbf{8} & \textbf{8} & \textbf{7} & \textbf{7} & \textbf{6} & \textbf{5} & \textbf{5} & \textcolor{gray}{3} & \textcolor{gray}{1} &  &  &  \\
DeepSeek V4.1 Flash & 0.68 & \textbf{16} & \textbf{15} & \textbf{15} & \textbf{15} & \textbf{14} & \textbf{12} & \textbf{11} & \textbf{11} & \textbf{10} & \textbf{10} & \textbf{9} & \textbf{9} & \textbf{9} & \textbf{8} & \textbf{8} & \textbf{7} & \textbf{7} & \textbf{6} & \textbf{5} & \textbf{5} & \textbf{3} & \textcolor{gray}{1} & \textcolor{gray}{0} &  &  \\
Qwen3.5-4B$^\S$ & 0.66 & \textbf{18} & \textbf{17} & \textbf{17} & \textbf{17} & \textbf{16} & \textbf{14} & \textbf{13} & \textbf{13} & \textbf{12} & \textbf{12} & \textbf{11} & \textbf{11} & \textbf{11} & \textbf{10} & \textbf{10} & \textbf{9} & \textbf{9} & \textbf{8} & \textbf{7} & \textbf{7} & \textbf{5} & \textcolor{gray}{3} & \textcolor{gray}{2} & \textcolor{gray}{2} &  \\
GPT-6 Luna & 0.62 & \textbf{22} & \textbf{21} & \textbf{21} & \textbf{21} & \textbf{20} & \textbf{18} & \textbf{17} & \textbf{17} & \textbf{16} & \textbf{16} & \textbf{15} & \textbf{15} & \textbf{15} & \textbf{14} & \textbf{14} & \textbf{13} & \textbf{13} & \textbf{12} & \textbf{11} & \textbf{11} & \textbf{9} & \textbf{7} & \textbf{6} & \textbf{6} & \textbf{4} \\
\bottomrule\end{tabular}
\end{table}

\begin{table}[t]\centering\footnotesize\caption{Prices behind the costs (USD per 1M tokens; thinking priced at the output rate), where the price was read and on which day. Every price is the list price of the model's developer on its own API, whichever host served our calls (Alibaba Cloud prices differ by deployment region; we use International, the only one outside China offered for all three Qwen models). DeepSeek is the only developer that prices by time of day, off-peak at half of peak; we quote the peak rate. $^\dagger$ Google lists a per-token price for Gemma 4 26B only; we use it for Gemma 4 31B as well.}\label{tab:static_v2_prices}
\begin{tabular}{@{}lrrll@{}}\toprule
Model & input & output & source & read \\ \midrule
Gemini 3.8 Flash & 0.75 & 3.75 & Google$^{a}$ & 2026-09-13 \\
Gemini 3.5 Flash & 0.75 & 3.75 & Google$^{a}$ & 2026-09-13 \\
Gemini 3.1 Flash Lite & 0.30 & 2.50 & Google$^{a}$ & 2026-09-13 \\
Gemma 4 31B$^\dagger$ & 0.15 & 0.60 & Google Cloud (Vertex AI)$^{b}$ & 2026-09-20 \\
Gemma 4 26B-A4B & 0.15 & 0.60 & Google Cloud (Vertex AI)$^{b}$ & 2026-09-20 \\
GPT-6 Astra & 10.00 & 50.00 & OpenAI$^{c}$ & 2026-09-15 \\
GPT-6 Sol & 2.00 & 10.00 & OpenAI$^{c}$ & 2026-09-23 \\
GPT-6 Luna & 0.10 & 0.50 & OpenAI$^{c}$ & 2026-09-23 \\
GPT-5.6 Sol & 4.00 & 20.00 & OpenAI$^{c}$ & 2026-09-15 \\
GPT-5.6 Terra & 2.00 & 12.00 & OpenAI$^{c}$ & 2026-09-15 \\
GPT-5.6 Luna & 0.20 & 1.20 & OpenAI$^{c}$ & 2026-09-15 \\
Claude Fable 5.1 & 10.00 & 50.00 & Anthropic$^{d}$ & 2026-09-15 \\
Claude Opus 5.5 & 4.00 & 20.00 & Anthropic$^{d}$ & 2026-09-23 \\
Claude Opus 5 & 5.00 & 25.00 & Anthropic$^{d}$ & 2026-09-15 \\
Claude Sonnet 5 & 2.00 & 10.00 & Anthropic$^{d}$ & 2026-09-15 \\
Qwen 3.8 Max & 2.00 & 6.00 & Alibaba Cloud (International)$^{e}$ & 2026-09-20 \\
Qwen 3.8 27B & 0.50 & 3.00 & Alibaba Cloud (International)$^{e}$ & 2026-09-20 \\
Qwen 3.5 397B-A17B & 0.60 & 3.60 & Alibaba Cloud (International)$^{e}$ & 2026-09-20 \\
Grok 4.6 & 2.00 & 6.00 & xAI$^{f}$ & 2026-09-15 \\
Kimi K3 & 3.00 & 15.00 & Moonshot AI$^{g}$ & 2026-09-20 \\
Muse Spark 1.3 & 1.25 & 4.25 & Meta$^{h}$ & 2026-09-20 \\
DeepSeek V4.1 Flash & 0.30 & 1.20 & DeepSeek$^{i}$ & 2026-09-20 \\
GLM-5.3 Flash & 0.15 & 0.50 & Z.AI$^{j}$ & 2026-09-20 \\
\midrule Gemma 4 E4B & -- & -- & no list price (served by us) & \\
Qwen3.5-4B & -- & -- & no list price (served by us) & \\
LFM2.5-VL-3B & -- & -- & no list price (served by us) & \\
\bottomrule\end{tabular}

\vspace{0.5em}{\scriptsize $^{a}$\,\url{https://ai.google.dev/gemini-api/docs/pricing} \quad $^{b}$\,\url{https://cloud.google.com/vertex-ai/generative-ai/pricing} \quad $^{c}$\,\url{https://developers.openai.com/api/docs/pricing} \quad $^{d}$\,\url{https://platform.claude.com/docs/en/about-claude/pricing} \quad $^{e}$\,\url{https://www.alibabacloud.com/help/en/model-studio/model-pricing} \quad $^{f}$\,\url{https://docs.x.ai/docs/models} \quad $^{g}$\,\url{https://platform.kimi.ai/docs/pricing/chat} \quad $^{h}$\,\url{https://dev.meta.ai/docs/pricing-rate-limits} \quad $^{i}$\,\url{https://api-docs.deepseek.com/quick_start/pricing} \quad $^{j}$\,\url{https://docs.z.ai/guides/overview/pricing}}
\end{table}

\begin{table}[t]\centering\scriptsize\caption{Acceptability F1 per environment: positives are the frames the scorer accepts; precision on the random frames, recall on the random frames plus the positive stratum.}\label{tab:static_v2_f1}
\begin{tabular}{@{}lrrrrrrrrrrr@{}}\toprule
Model & breaker & plate & car view & damage & equip. & horizon & rail & RO board & slope & tape & wire \\ \midrule
Gemini 3.8 Flash & 0.45 & -- & 0.38 & 0.14 & 0.41 & 0.54 & 0.45 & 0.73 & 0.62 & 0.40 & 0.24 \\
Gemini 3.5 Flash & 0.41 & -- & 0.53 & 0.13 & 0.39 & 0.65 & 0.60 & 0.59 & 0.71 & 0.21 & 0.47 \\
Gemini 3.1 Flash Lite & 0.34 & -- & 0.49 & 0.14 & 0.28 & 0.55 & 0.31 & 0.62 & 0.37 & 0.37 & 0.43 \\
Gemma 4 31B & 0.46 & -- & 0.39 & 0.08 & 0.37 & 0.60 & 0.36 & 0.66 & 0.57 & 0.16 & 0.48 \\
Gemma 4 26B-A4B & 0.38 & -- & 0.25 & 0.25 & 0.35 & 0.54 & 0.55 & 0.59 & 0.50 & 0.20 & 0.34 \\
GPT-6 Astra & 0.66 & -- & 0.61 & 0.26 & 0.62 & 0.63 & 0.48 & 0.73 & 0.63 & 0.72 & 0.55 \\
GPT-6 Sol & 0.59 & -- & 0.60 & 0.34 & 0.56 & 0.63 & 0.44 & 0.66 & 0.64 & 0.37 & 0.45 \\
GPT-6 Luna & 0.40 & -- & 0.47 & 0.32 & 0.60 & 0.30 & 0.35 & 0.65 & 0.26 & 0.40 & 0.30 \\
GPT-5.6 Sol & 0.78 & -- & 0.58 & 0.41 & 0.45 & 0.61 & 0.48 & 0.71 & 0.59 & 0.45 & 0.54 \\
GPT-5.6 Terra & 0.58 & -- & 0.55 & 0.24 & 0.46 & 0.62 & 0.47 & 0.68 & 0.53 & 0.38 & 0.48 \\
GPT-5.6 Luna & 0.59 & -- & 0.53 & 0.30 & 0.45 & 0.55 & 0.39 & 0.71 & 0.68 & 0.17 & 0.49 \\
Claude Fable 5.1 & 0.59 & -- & 0.65 & 0.38 & 0.55 & 0.66 & 0.24 & 0.71 & 0.67 & 0.28 & 0.35 \\
Claude Opus 5.5 & 0.55 & -- & 0.65 & 0.27 & 0.47 & 0.70 & 0.44 & 0.64 & 0.71 & 0.34 & 0.56 \\
Claude Opus 5 & 0.54 & -- & 0.60 & 0.38 & 0.57 & 0.67 & 0.32 & 0.73 & 0.49 & 0.37 & 0.48 \\
Claude Sonnet 5 & 0.58 & -- & 0.48 & 0.27 & 0.55 & 0.52 & 0.31 & 0.76 & 0.23 & -- & 0.37 \\
Qwen 3.8 Max & 0.38 & -- & 0.63 & 0.29 & 0.34 & 0.62 & 0.46 & 0.67 & 0.64 & 0.19 & 0.63 \\
Qwen 3.8 27B & 0.47 & -- & 0.46 & 0.47 & 0.38 & 0.61 & 0.49 & 0.71 & 0.60 & 0.12 & 0.34 \\
Qwen 3.5 397B-A17B & 0.45 & -- & 0.14 & 0.27 & 0.41 & 0.65 & 0.49 & 0.72 & 0.70 & -- & 0.29 \\
Grok 4.6 & 0.41 & -- & 0.21 & 0.34 & 0.48 & 0.64 & 0.48 & 0.64 & 0.52 & 0.37 & 0.60 \\
Kimi K3 & 0.52 & -- & 0.50 & 0.28 & 0.42 & 0.63 & 0.41 & 0.71 & 0.60 & 0.00 & 0.57 \\
Muse Spark 1.3 & 0.54 & -- & 0.51 & 0.26 & 0.47 & 0.56 & 0.33 & 0.66 & 0.65 & 0.26 & 0.18 \\
DeepSeek V4.1 Flash & 0.30 & -- & 0.41 & 0.40 & 0.34 & 0.60 & 0.46 & 0.67 & 0.51 & 0.18 & 0.09 \\
GLM-5.3 Flash & 0.64 & -- & 0.51 & 0.46 & 0.38 & 0.62 & 0.32 & 0.73 & 0.63 & 0.37 & 0.49 \\
\midrule Gemma 4 E4B & 0.26 & -- & 0.46 & 0.40 & 0.35 & 0.47 & 0.29 & 0.61 & 0.45 & 0.24 & 0.36 \\
Qwen3.5-4B & 0.40 & -- & 0.23 & 0.30 & 0.38 & 0.43 & 0.18 & 0.65 & 0.33 & -- & 0.34 \\
LFM2.5-VL-3B & 0.23 & -- & 0.57 & 0.53 & 0.26 & -- & 0.24 & 0.45 & 0.56 & 0.47 & 0.44 \\
\bottomrule\end{tabular}\end{table}

\begin{table}[t]\centering\scriptsize\caption{False-accept rate (\%): frames the scorer rejects that the model accepts, on the 100 random field-pose frames per environment.}\label{tab:static_v2_fa}
\begin{tabular}{@{}lrrrrrrrrrrr@{}}\toprule
Model & breaker & plate & car view & damage & equip. & horizon & rail & RO board & slope & tape & wire \\ \midrule
\emph{scorer-acceptable (\%)} & 5 & 21 & 26 & 28 & 6 & 19 & 12 & 18 & 27 & 18 & 39 \\ \midrule
Gemini 3.8 Flash & 9 & -- & 5 & 5 & 18 & 11 & 5 & 15 & 33 & 3 & 7 \\
Gemini 3.5 Flash & 15 & -- & 8 & 6 & 20 & 12 & 3 & 29 & 20 & 1 & 11 \\
Gemini 3.1 Flash Lite & 20 & -- & 12 & 3 & 32 & 21 & 11 & 26 & 8 & 4 & 20 \\
Gemma 4 31B & 9 & -- & 16 & 2 & 21 & 22 & 3 & 22 & 33 & 3 & 20 \\
Gemma 4 26B-A4B & 17 & -- & 8 & 18 & 23 & 26 & 8 & 29 & 33 & 1 & 8 \\
GPT-6 Astra & 0 & -- & 17 & 15 & 5 & 15 & 7 & 8 & 41 & 5 & 23 \\
GPT-6 Sol & 0 & -- & 16 & 39 & 10 & 19 & 0 & 11 & 36 & 4 & 10 \\
GPT-6 Luna & 1 & -- & 9 & 11 & 6 & 20 & 5 & 6 & 12 & 8 & 10 \\
GPT-5.6 Sol & 0 & -- & 17 & 24 & 13 & 16 & 3 & 12 & 50 & 12 & 21 \\
GPT-5.6 Terra & 6 & -- & 14 & 21 & 15 & 21 & 8 & 14 & 55 & 5 & 18 \\
GPT-5.6 Luna & 2 & -- & 30 & 24 & 13 & 19 & 14 & 11 & 20 & 3 & 33 \\
Claude Fable 5.1 & 2 & -- & 16 & 24 & 7 & 11 & 9 & 15 & 30 & 3 & 13 \\
Claude Opus 5.5 & 7 & -- & 19 & 23 & 11 & 12 & 19 & 12 & 9 & 4 & 25 \\
Claude Opus 5 & 5 & -- & 11 & 16 & 9 & 14 & 15 & 14 & 5 & 5 & 15 \\
Claude Sonnet 5 & 6 & -- & 14 & 13 & 10 & 22 & 25 & 12 & 5 & 0 & 8 \\
Qwen 3.8 Max & 17 & -- & 22 & 27 & 24 & 21 & 18 & 20 & 42 & 0 & 48 \\
Qwen 3.8 27B & 12 & -- & 11 & 19 & 20 & 22 & 8 & 15 & 28 & 0 & 15 \\
Qwen 3.5 397B-A17B & 13 & -- & 5 & 3 & 18 & 16 & 2 & 15 & 18 & 0 & 15 \\
Grok 4.6 & 15 & -- & 3 & 24 & 14 & 22 & 7 & 23 & 50 & 4 & 34 \\
Kimi K3 & 3 & -- & 25 & 40 & 17 & 21 & 18 & 15 & 47 & 1 & 23 \\
Muse Spark 1.3 & 6 & -- & 9 & 5 & 12 & 17 & 3 & 14 & 38 & 1 & 2 \\
DeepSeek V4.1 Flash & 19 & -- & 12 & 6 & 21 & 20 & 17 & 20 & 45 & 0 & 7 \\
GLM-5.3 Flash & 5 & -- & 9 & 32 & 17 & 19 & 2 & 14 & 29 & 4 & 16 \\
\midrule Gemma 4 E4B & 29 & -- & 53 & 52 & 22 & 35 & 27 & 28 & 61 & 17 & 18 \\
Qwen3.5-4B & 16 & -- & 0 & 16 & 21 & 1 & 70 & 20 & 15 & 0 & 8 \\
LFM2.5-VL-3B & 36 & -- & 39 & 50 & 37 & 0 & 69 & 52 & 59 & 49 & 5 \\
\bottomrule\end{tabular}\end{table}

\begin{sidewaystable}[p]\centering\scriptsize\setlength{\tabcolsep}{3.5pt}\caption{Documentation score by field and model on scorer-acceptable frames.}\label{tab:field_scores_full}
\begin{tabular}{@{}lrrrrrrrrrrrrrrrrrrrrrrrrrr@{}}\toprule
Field & \rotatebox{90}{Gemini 3.8 Flash} & \rotatebox{90}{Gemini 3.5 Flash} & \rotatebox{90}{Gemini 3.1 Flash Lite} & \rotatebox{90}{Gemma 4 31B} & \rotatebox{90}{Gemma 4 26B-A4B} & \rotatebox{90}{Gemma 4 E4B} & \rotatebox{90}{GPT-6 Astra} & \rotatebox{90}{GPT-6 Sol} & \rotatebox{90}{GPT-6 Luna} & \rotatebox{90}{GPT-5.6 Sol} & \rotatebox{90}{GPT-5.6 Terra} & \rotatebox{90}{GPT-5.6 Luna} & \rotatebox{90}{Claude Fable 5.1} & \rotatebox{90}{Claude Opus 5.5} & \rotatebox{90}{Claude Opus 5} & \rotatebox{90}{Claude Sonnet 5} & \rotatebox{90}{Qwen 3.8 Max} & \rotatebox{90}{Qwen 3.8 27B} & \rotatebox{90}{Qwen 3.5 397B-A17B} & \rotatebox{90}{Qwen3.5-4B} & \rotatebox{90}{Grok 4.6} & \rotatebox{90}{Kimi K3} & \rotatebox{90}{Muse Spark 1.3} & \rotatebox{90}{DeepSeek V4.1 Flash} & \rotatebox{90}{GLM-5.3 Flash} & \rotatebox{90}{LFM2.5-VL-3B} \\ \midrule
breaker: main amperage & 0.98 & 0.98 & 0.98 & 0.98 & 0.98 & 0.98 & 1.00 & 0.98 & 1.00 & 0.98 & 0.98 & 0.98 & 0.98 & 0.98 & 0.98 & 0.98 & 1.00 & 0.98 & 0.93 & 0.98 & 0.98 & 1.00 & 0.98 & 1.00 & 0.98 & 0.98 \\
breaker: circuit rows (fraction right) & 0.99 & 0.93 & 0.98 & 0.96 & 0.97 & 0.68 & 0.98 & 0.95 & 0.81 & 0.92 & 0.86 & 0.87 & 0.96 & 0.98 & 0.95 & 0.89 & 0.96 & 0.86 & 0.88 & 0.94 & 0.92 & 0.92 & 0.96 & 0.80 & 0.84 & 0.67 \\
plate: characters & 0.95 & 0.95 & 1.00 & 1.00 & 0.95 & 0.95 & 1.00 & 0.95 & 0.48 & 0.90 & 0.90 & 0.95 & 1.00 & 1.00 & 1.00 & 1.00 & 1.00 & 0.95 & 1.00 & 1.00 & 1.00 & 1.00 & 1.00 & 1.00 & 1.00 & 1.00 \\
plate: state & 1.00 & 1.00 & 1.00 & 1.00 & 1.00 & 1.00 & 1.00 & 1.00 & 1.00 & 1.00 & 1.00 & 1.00 & 1.00 & 1.00 & 1.00 & 1.00 & 1.00 & 1.00 & 1.00 & 1.00 & 1.00 & 1.00 & 1.00 & 1.00 & 1.00 & 1.00 \\
label: brand & 1.00 & 1.00 & 1.00 & 1.00 & 1.00 & 0.96 & 1.00 & 1.00 & 1.00 & 0.98 & 1.00 & 1.00 & 1.00 & 1.00 & 1.00 & 1.00 & 1.00 & 1.00 & 1.00 & 0.89 & 1.00 & 1.00 & 1.00 & 0.96 & 1.00 & 1.00 \\
label: model number & 1.00 & 0.96 & 1.00 & 1.00 & 0.98 & 0.91 & 0.98 & 0.84 & 0.75 & 0.89 & 0.79 & 0.93 & 1.00 & 1.00 & 0.98 & 0.96 & 0.96 & 0.93 & 1.00 & 0.95 & 0.98 & 0.98 & 0.98 & 0.95 & 0.96 & 0.88 \\
label: serial number & 1.00 & 0.95 & 1.00 & 0.98 & 1.00 & 0.89 & 0.93 & 0.93 & 0.38 & 0.89 & 0.93 & 0.79 & 0.95 & 0.98 & 1.00 & 0.93 & 0.96 & 0.96 & 0.91 & 0.95 & 0.96 & 1.00 & 0.96 & 0.93 & 0.89 & 0.50 \\
RO board: RO number & 0.98 & 0.98 & 0.98 & 0.96 & 0.98 & 0.86 & 0.98 & 0.88 & 0.70 & 0.89 & 0.95 & 0.89 & 0.98 & 0.98 & 0.98 & 0.98 & 0.96 & 0.88 & 0.93 & 0.95 & 0.96 & 0.98 & 0.96 & 0.96 & 0.86 & 0.96 \\
RO board: VIN & 0.89 & 0.82 & 0.80 & 0.73 & 0.73 & 0.46 & 0.46 & 0.32 & 0.12 & 0.39 & 0.43 & 0.27 & 0.88 & 0.86 & 0.88 & 0.86 & 0.59 & 0.45 & 0.61 & 0.52 & 0.61 & 0.73 & 0.59 & 0.43 & 0.45 & 0.41 \\
gauge: display reading ($\pm0.2^\circ$) & 0.95 & 0.95 & 0.95 & 0.91 & 0.88 & 0.57 & 0.89 & 0.55 & 0.08 & 0.57 & 0.41 & 0.28 & 0.83 & 0.83 & 0.53 & 0.63 & 0.95 & 0.51 & 0.81 & 0.69 & 0.52 & 0.59 & 0.64 & 0.41 & 0.52 & 0.60 \\
\midrule
car view: face shown & 0.96 & 0.83 & 0.78 & 0.74 & 0.78 & 0.57 & 0.96 & 0.87 & 0.70 & 0.74 & 0.87 & 0.74 & 0.78 & 1.00 & 0.78 & 0.83 & 1.00 & 0.83 & 0.70 & 0.83 & 0.96 & 0.91 & 1.00 & 0.87 & 1.00 & 0.70 \\
car view: body style & 0.87 & 0.91 & 0.83 & 0.74 & 0.74 & 0.39 & 0.87 & 0.87 & 0.52 & 0.87 & 0.74 & 0.78 & 0.96 & 1.00 & 0.87 & 0.96 & 0.74 & 0.83 & 0.78 & 0.87 & 0.87 & 0.87 & 0.91 & 0.74 & 0.91 & 0.74 \\
damage: kinds & 0.92 & 0.92 & 0.83 & 0.83 & 0.83 & 0.67 & 0.92 & 0.92 & 0.92 & 0.92 & 0.96 & 0.96 & 1.00 & 0.92 & 0.96 & 0.88 & 0.92 & 0.88 & 0.96 & 0.83 & 0.92 & 0.92 & 0.83 & 0.92 & 0.88 & 0.75 \\
damage: severity & 0.42 & 0.67 & 0.67 & 0.75 & 0.71 & 0.71 & 0.42 & 0.58 & 0.67 & 0.58 & 0.67 & 0.62 & 0.62 & 0.46 & 0.62 & 0.71 & 0.50 & 0.67 & 0.62 & 0.33 & 0.62 & 0.67 & 0.67 & 0.75 & 0.75 & 0.50 \\
label: what it is on & 1.00 & 1.00 & 1.00 & 1.00 & 0.98 & 0.98 & 1.00 & 1.00 & 0.98 & 1.00 & 1.00 & 1.00 & 1.00 & 1.00 & 1.00 & 1.00 & 1.00 & 1.00 & 1.00 & 0.57 & 1.00 & 1.00 & 1.00 & 0.98 & 1.00 & 0.20 \\
horizon: visible & 0.99 & 0.99 & 0.91 & 0.99 & 0.99 & 0.94 & 0.99 & 1.00 & 1.00 & 0.99 & 1.00 & 0.99 & 1.00 & 0.99 & 0.99 & 0.99 & 1.00 & 1.00 & 1.00 & 0.99 & 0.97 & 1.00 & 0.99 & 0.91 & 0.97 & 0.84 \\
horizon: level & 0.96 & 0.96 & 0.94 & 0.97 & 0.80 & 0.93 & 0.93 & 0.99 & 0.88 & 0.91 & 0.96 & 0.93 & 0.99 & 0.99 & 0.97 & 0.94 & 0.97 & 0.94 & 0.90 & 0.68 & 0.96 & 0.97 & 0.97 & 0.88 & 0.94 & 0.42 \\
rail: lag bolt present & 0.76 & 0.61 & 0.50 & 0.61 & 0.68 & 0.47 & 0.77 & 0.61 & 0.71 & 0.77 & 0.76 & 0.65 & 0.74 & 0.74 & 0.60 & 0.61 & 0.82 & 0.68 & 0.69 & 0.40 & 0.73 & 0.73 & 0.53 & 0.71 & 0.61 & 0.79 \\
rail: flashing present & 0.95 & 0.74 & 0.37 & 0.81 & 0.84 & 0.69 & 1.00 & 1.00 & 0.79 & 1.00 & 0.97 & 0.97 & 1.00 & 0.97 & 0.81 & 0.92 & 1.00 & 0.89 & 0.85 & 0.76 & 0.97 & 0.89 & 0.94 & 0.94 & 0.97 & 0.97 \\
gauge: on module vs bare roof & 0.97 & 0.93 & 0.27 & 0.81 & 0.76 & 0.84 & 0.97 & 1.00 & 0.40 & 1.00 & 0.95 & 0.96 & 0.97 & 0.79 & 0.52 & 0.23 & 1.00 & 0.80 & 0.87 & 0.37 & 0.92 & 0.96 & 1.00 & 0.79 & 0.93 & 1.00 \\
wire: touching roof & 0.67 & 0.64 & 0.64 & 0.74 & 0.59 & 0.64 & 0.51 & 0.54 & 0.59 & 0.51 & 0.51 & 0.44 & 0.59 & 0.59 & 0.49 & 0.46 & 0.64 & 0.64 & 0.54 & 0.38 & 0.62 & 0.49 & 0.62 & 0.62 & 0.59 & 0.64 \\
tape: leaning & 0.60 & 0.54 & 0.57 & 0.52 & 0.52 & 0.52 & 0.82 & 0.59 & 0.57 & 0.62 & 0.60 & 0.48 & 0.57 & 0.60 & 0.59 & 0.48 & 0.52 & 0.48 & 0.48 & 0.47 & 0.59 & 0.45 & 0.57 & 0.29 & 0.55 & 0.76 \\
\midrule
tape: number at the crossing ($\pm\nicefrac{1}{8}$\,in) & 0.18 & 0.12 & 0.04 & 0.06 & 0.04 & 0.03 & 0.30 & 0.36 & 0.07 & 0.27 & 0.18 & 0.12 & 0.13 & 0.36 & 0.15 & 0.01 & 0.07 & 0.06 & 0.01 & 0.01 & 0.01 & 0.03 & 0.04 & 0.01 & 0.04 & 0.07 \\
tape: true gap ($\pm\nicefrac{1}{4}$\,in) & 0.33 & 0.30 & 0.16 & 0.13 & 0.10 & 0.06 & 0.34 & 0.43 & 0.18 & 0.40 & 0.31 & 0.24 & 0.34 & 0.48 & 0.24 & 0.10 & 0.34 & 0.24 & 0.19 & 0.12 & 0.16 & 0.24 & 0.24 & 0.07 & 0.13 & 0.04 \\
\bottomrule\end{tabular}\end{sidewaystable}

\begin{table}[t]\centering\scriptsize\setlength{\tabcolsep}{4pt}\caption{Token price against set cost, models with a list price, sorted by output price. Prompt, visible output and thinking tokens are per item (means over the 1,378 items; Anthropic reports one output count, so thinking is that count minus the tokens of the reply we received, counted with the provider's tokenizer). \emph{Prompt share} is the fraction of the set's cost that is prompt tokens; the rest is output and thinking at the output rate. The last two columns rank the models by output price and by the cost of the set.}\label{tab:cost_split}
\begin{tabular}{@{}lrrrrrrrrr@{}}\toprule
Model & in \$/M & out \$/M & prompt & output & thinking & prompt share & set (\$) & rank by price & rank by set \\ \midrule
GLM-5.3 Flash & 0.15 & 0.50 & 1,289 & 37 & 1,460 & 21\% & 1.30 & 1 & 6 \\
GPT-6 Luna & 0.10 & 0.50 & 1,020 & 41 & 751 & 20\% & 0.69 & 2 & 1 \\
Gemma 4 26B-A4B & 0.15 & 0.60 & 527 & 43 & 1,325 & 9\% & 1.24 & 3 & 5 \\
Gemma 4 31B & 0.15 & 0.60 & 527 & 43 & 677 & 15\% & 0.70 & 4 & 2 \\
DeepSeek V4.1 Flash & 0.30 & 1.20 & 786 & 143 & 4,903 & 4\% & 8.67 & 5 & 12 \\
GPT-5.6 Luna & 0.20 & 1.20 & 1,170 & 39 & 482 & 27\% & 1.18 & 6 & 4 \\
Gemini 3.1 Flash Lite & 0.30 & 2.50 & 1,333 & 52 & 0 & 75\% & 0.73 & 7 & 3 \\
Qwen 3.8 27B & 0.50 & 3.00 & 1,122 & 39 & 2,763 & 6\% & 12.36 & 8 & 13 \\
Qwen 3.5 397B-A17B & 0.60 & 3.60 & 1,027 & 60 & 2,799 & 6\% & 15.03 & 9 & 15 \\
Gemini 3.5 Flash & 0.75 & 3.75 & 1,333 & 43 & 1,114 & 19\% & 7.36 & 10 & 9 \\
Gemini 3.8 Flash & 0.75 & 3.75 & 1,333 & 36 & 1,083 & 19\% & 7.16 & 11 & 8 \\
Muse Spark 1.3 & 1.25 & 4.25 & 1,221 & 74 & 1,825 & 16\% & 13.22 & 12 & 14 \\
Grok 4.6 & 2.00 & 6.00 & 1,023 & 35 & 2,634 & 11\% & 24.88 & 13 & 19 \\
Qwen 3.8 Max & 2.00 & 6.00 & 1,040 & 41 & 4,231 & 8\% & 38.19 & 14 & 21 \\
Claude Sonnet 5 & 2.00 & 10.00 & 1,406 & 48 & 19 & 81\% & 4.80 & 15 & 7 \\
GPT-6 Sol & 2.00 & 10.00 & 1,170 & 36 & 320 & 40\% & 8.13 & 16 & 10 \\
GPT-5.6 Terra & 2.00 & 12.00 & 1,170 & 37 & 282 & 38\% & 8.49 & 17 & 11 \\
Kimi K3 & 3.00 & 15.00 & 1,370 & 46 & 1,582 & 14\% & 39.31 & 18 & 22 \\
Claude Opus 5.5 & 4.00 & 20.00 & 1,408 & 50 & 263 & 47\% & 16.36 & 19 & 16 \\
GPT-5.6 Sol & 4.00 & 20.00 & 1,170 & 38 & 368 & 37\% & 17.65 & 20 & 17 \\
Claude Opus 5 & 5.00 & 25.00 & 1,406 & 50 & 345 & 42\% & 23.32 & 21 & 18 \\
Claude Fable 5.1 & 10.00 & 50.00 & 1,408 & 52 & 267 & 47\% & 41.35 & 22 & 23 \\
GPT-6 Astra & 10.00 & 50.00 & 1,170 & 37 & 165 & 54\% & 30.05 & 23 & 20 \\
\bottomrule\end{tabular}\end{table}

\begin{table}[t]\centering\scriptsize\caption{Misreading and rejection on scorer-acceptable frames, where the property the requirement checks always holds, so the right verdict is accept and reporting the defect is a misreading. Pooled over the 26 models: model answers, how often the defect is reported, how often a report comes with a rejection in the same reply, how many frames are rejected, and what share of those rejections come with the defect reported. \env{rail\_attachment} frames with a missing part are not scored for acceptability.}\label{tab:accept_by_answer}
\begin{tabular}{@{}llrrrrr@{}}\toprule
Environment & Defect reported & answers & reports defect & rejects if reported & rejected & of those, defect reported \\ \midrule
\env{tape\_gap} & tape is tilted & 1,688 & 75\% & 99.6\% & 1,296 & 97\% \\
\env{slope\_gauge} & gauge is not on the module & 1,950 & 19\% & 96.0\% & 488 & 74\% \\
\env{rail\_attachment} & lag bolt or flashing missing & 1,273 & 35\% & 88.8\% & 626 & 63\% \\
\bottomrule\end{tabular}\end{table}

\FloatBarrier

\section{Human ratings}
\label{app:human}
Table~\ref{tab:human_ratings} gives, per environment, the random frames the raters judged, their agreement with each other, and the frozen scorer's agreement with their majority on the held-out frames (Section~\ref{sec:metrics}).
\begin{table}[h]\centering\scriptsize\caption{Human acceptability ratings. \emph{Frames}: random frames with at least two ratings (the frames used to adjust the scorer and the held-out frames); \emph{agree}: raw agreement over all rater pairs on a frame; $\alpha$: Krippendorff's alpha (nominal), which corrects for chance and allows raters to skip frames. \emph{Scorer}: the frozen scorer's agreement with the raters' majority on the held-out frames with at least three ratings, ties excluded (all environments: 82\% of 95). $^\dagger$ acceptability not scored (the raters split on the requirement); frames the frozen scorer leaves unscored are not in the scorer column.}\label{tab:human_ratings}
\begin{tabular}{@{}lrrrrrr@{}}\toprule
Set & Frames & Raters & Ratings/frame & Agree & $\alpha$ & Scorer (held-out) \\ \midrule
\env{breaker\_label} & 30 & 5 & 4.1 & 0.94 & 0.79 & 0.91 (11) \\
\env{equipment\_label} & 29 & 5 & 4.3 & 0.90 & 0.58 & 0.91 (11) \\
\env{ro\_board} & 30 & 5 & 4.1 & 0.98 & 0.86 & 1.00 (8) \\
\env{rail\_attachment} & 30 & 5 & 4.1 & 0.88 & 0.53 & 0.80 (5) \\
\env{car\_plate}$^\dagger$ & 28 & 5 & 4.2 & 0.80 & 0.54 & -- \\
\env{wire\_adequacy} & 30 & 5 & 4.3 & 0.78 & 0.56 & 0.91 (11) \\
\env{tape\_gap} & 30 & 6 & 5.1 & 0.72 & 0.44 & 0.73 (11) \\
\env{slope\_gauge} & 30 & 5 & 4.1 & 0.73 & 0.46 & 0.80 (10) \\
\env{car\_view} & 28 & 5 & 4.2 & 0.71 & 0.49 & 0.60 (10) \\
\env{horizon\_peak} & 30 & 5 & 4.0 & 0.65 & 0.25 & 0.89 (9) \\
\env{damage\_view} & 30 & 6 & 5.0 & 0.64 & 0.19 & 0.67 (9) \\
\bottomrule\end{tabular}\end{table}

\FloatBarrier

\section{Real-photo details}
\label{app:real}
Table~\ref{tab:real_vs_sim} gives the per-model numbers behind Figure~\ref{fig:static_cost} (bottom), with intervals.
\begin{table}[t]\centering\scriptsize\setlength{\tabcolsep}{3pt}\caption{Synthetic against real, per model, three matched questions. Synthetic: the share of scorer-acceptable synthetic frames whose field is right (\env{tape\_gap} reading within $\nicefrac{1}{8}$\,in; \env{car\_view} face; \env{damage\_view} severity on the frames where the scorer finds the damage readable: sharp, visible and resolved, framing not required). Real: our tape photos read within $\nicefrac{1}{8}$\,in (133 readings on 84 photos of 35 items); the face named is one the labels list as visible (117 side-determinable frames); severity on the 59 real damaged-panel photos an author rated on the prompt's area scale after the model runs, blind to the models' answers (4 rated can't tell are left out). Brackets: 95\% bootstrap intervals over frames, or over items for the tape photos. Across the 22 models, synthetic against real (Spearman, Pearson): tape $\rho=0.68$ ($p=0.001$), $r=0.82$ ($p<0.001$); face $\rho=0.79$ ($p<0.001$), $r=0.85$ ($p<0.001$); severity $\rho=0.63$ ($p=0.002$), $r=0.58$ ($p=0.005$). Not run on the real photos: Qwen 3.8 Max, Muse Spark 1.3, Kimi K3, DeepSeek V4.1 Flash. Tape replies cut by the output cap, left out: GLM-5.3 Flash 54, Qwen 3.8 27B 15, Qwen 3.5 397B-A17B 8.}\label{tab:real_vs_sim}
\begin{tabular}{@{}lrrrrrr@{}}\toprule
Model & tape: synthetic & tape: real & face: synthetic & face: real & severity: synthetic & severity: real \\ \midrule
Claude Opus 5.5 & 0.36 [0.24, 0.48] & 0.35 [0.26, 0.47] & 1.00 [1.00, 1.00] & 0.96 [0.91, 0.99] & 0.63 [0.51, 0.73] & 0.56 [0.44, 0.68] \\
GPT-6 Sol & 0.36 [0.24, 0.48] & 0.33 [0.25, 0.41] & 0.87 [0.74, 1.00] & 0.89 [0.83, 0.94] & 0.51 [0.39, 0.63] & 0.47 [0.36, 0.61] \\
GPT-6 Astra & 0.30 [0.19, 0.40] & 0.44 [0.34, 0.53] & 0.96 [0.87, 1.00] & 0.95 [0.91, 0.98] & 0.42 [0.30, 0.54] & 0.42 [0.31, 0.54] \\
GPT-5.6 Sol & 0.27 [0.16, 0.37] & 0.33 [0.25, 0.41] & 0.74 [0.57, 0.91] & 0.79 [0.72, 0.86] & 0.46 [0.34, 0.58] & 0.42 [0.31, 0.54] \\
Gemini 3.8 Flash & 0.18 [0.09, 0.27] & 0.21 [0.13, 0.29] & 0.96 [0.87, 1.00] & 0.91 [0.85, 0.96] & 0.37 [0.27, 0.49] & 0.56 [0.42, 0.68] \\
GPT-5.6 Terra & 0.18 [0.09, 0.27] & 0.10 [0.05, 0.16] & 0.87 [0.70, 1.00] & 0.78 [0.70, 0.85] & 0.54 [0.42, 0.66] & 0.54 [0.41, 0.68] \\
Claude Opus 5 & 0.15 [0.07, 0.24] & 0.23 [0.15, 0.31] & 0.78 [0.61, 0.91] & 0.85 [0.79, 0.91] & 0.54 [0.42, 0.66] & 0.54 [0.41, 0.66] \\
Claude Fable 5.1 & 0.13 [0.06, 0.22] & 0.29 [0.20, 0.37] & 0.78 [0.61, 0.96] & 0.83 [0.76, 0.90] & 0.61 [0.49, 0.73] & 0.59 [0.46, 0.71] \\
Gemini 3.5 Flash & 0.12 [0.04, 0.21] & 0.17 [0.10, 0.24] & 0.83 [0.65, 0.96] & 0.88 [0.82, 0.94] & 0.69 [0.58, 0.79] & 0.69 [0.58, 0.81] \\
GPT-5.6 Luna & 0.12 [0.04, 0.19] & 0.19 [0.13, 0.26] & 0.74 [0.57, 0.91] & 0.81 [0.74, 0.88] & 0.43 [0.31, 0.54] & 0.39 [0.27, 0.51] \\
GPT-6 Luna & 0.07 [0.01, 0.13] & 0.11 [0.05, 0.17] & 0.70 [0.52, 0.87] & 0.77 [0.69, 0.84] & 0.64 [0.52, 0.76] & 0.36 [0.24, 0.47] \\
LFM2.5-VL-3B & 0.07 [0.01, 0.15] & 0.03 [0.01, 0.06] & 0.70 [0.52, 0.87] & 0.75 [0.67, 0.83] & 0.42 [0.30, 0.54] & 0.47 [0.36, 0.59] \\
Gemma 4 31B & 0.06 [0.01, 0.12] & 0.11 [0.06, 0.16] & 0.74 [0.57, 0.91] & 0.73 [0.64, 0.80] & 0.70 [0.60, 0.81] & 0.59 [0.47, 0.71] \\
Qwen 3.8 27B & 0.06 [0.01, 0.12] & 0.13 [0.07, 0.19] & 0.83 [0.65, 0.96] & 0.85 [0.79, 0.91] & 0.61 [0.51, 0.72] & 0.53 [0.39, 0.66] \\
Gemini 3.1 Flash Lite & 0.04 [0.00, 0.10] & 0.11 [0.05, 0.17] & 0.78 [0.61, 0.91] & 0.79 [0.72, 0.86] & 0.67 [0.55, 0.78] & 0.64 [0.53, 0.76] \\
Gemma 4 26B-A4B & 0.04 [0.00, 0.10] & 0.11 [0.06, 0.19] & 0.78 [0.61, 0.96] & 0.81 [0.74, 0.88] & 0.64 [0.52, 0.76] & 0.59 [0.47, 0.71] \\
GLM-5.3 Flash & 0.04 [0.00, 0.10] & 0.27 [0.19, 0.35] & 1.00 [1.00, 1.00] & 0.92 [0.87, 0.97] & 0.60 [0.48, 0.72] & 0.58 [0.46, 0.71] \\
Gemma 4 E4B & 0.03 [0.00, 0.07] & 0.05 [0.02, 0.10] & 0.57 [0.35, 0.74] & 0.68 [0.59, 0.77] & 0.66 [0.54, 0.76] & 0.46 [0.32, 0.58] \\
Claude Sonnet 5 & 0.01 [0.00, 0.04] & 0.08 [0.04, 0.14] & 0.83 [0.65, 0.96] & 0.77 [0.69, 0.84] & 0.70 [0.58, 0.81] & 0.59 [0.47, 0.71] \\
Grok 4.6 & 0.01 [0.00, 0.04] & 0.13 [0.08, 0.19] & 0.96 [0.87, 1.00] & 0.84 [0.77, 0.90] & 0.69 [0.57, 0.79] & 0.56 [0.42, 0.68] \\
Qwen 3.5 397B-A17B & 0.01 [0.00, 0.04] & 0.13 [0.08, 0.18] & 0.70 [0.52, 0.87] & 0.78 [0.70, 0.85] & 0.67 [0.55, 0.78] & 0.58 [0.44, 0.71] \\
Qwen3.5-4B & 0.01 [0.00, 0.04] & 0.02 [0.00, 0.04] & 0.83 [0.65, 0.96] & 0.79 [0.72, 0.87] & 0.45 [0.33, 0.57] & 0.41 [0.29, 0.53] \\
\bottomrule\end{tabular}\end{table}

\end{document}